\documentclass[11pt,onecolumn,draftclsnofoot]{IEEEtran}
\usepackage{amsfonts}
\IEEEoverridecommandlockouts

\ifCLASSINFOpdf
\else
\fi
\usepackage{multirow}
\usepackage{epsfig}
\usepackage{graphicx}
\usepackage{subfigure}
\usepackage{makecell}
\usepackage{psfig}
\usepackage{epsf}
\usepackage[cmex10]{amsmath}
\usepackage{booktabs}
\usepackage{fancyhdr}
\usepackage[ruled,linesnumbered]{algorithm2e}
\usepackage{caption2}
\usepackage{array}
\usepackage{color}
\usepackage{cite}

\begin{document}
\title{Cognitive Semantic Communication Systems Driven by Knowledge Graph: Principle, Implementation, and Performance Evaluation}
\author{Fuhui Zhou, \emph{Senior Member, IEEE}, Yihao Li, Ming Xu, Lu Yuan,\\ Qihui Wu, \emph{Senior Member, IEEE}, Rose Qingyang Hu, \emph{Fellow, IEEE}, \\and Naofal Al-Dhahir, \emph{Fellow, IEEE}
\thanks{This work were presented in part at the IEEE International Conference on Communications (ICC) \cite{zhou2022cognitive}, Seoul, South Korea, May 2022.}
\thanks{F. Zhou, Y. Li, M. Xu, L. Yuan, and Q. Wu are with The College of Electronic and Information Engineering, Nanjing University of Astronautics and Aeronautics, P. R. China, 330031 (e-mail: zhoufuhui@nuaa.edu.cn; LiYiHao1999@nuaa.edu.cn; xuming98@nuaa.edu.cn; yuanlu@nuaa.edu.cn; wuqihui2014@sina.com).}
\thanks{R. Q. Hu is with the Department of Electrical and Computer Engineering, Utah State University, Logan, UT 84322 USA (e-mail: rose.hu@usu.edu).}
\thanks{N. Al-Dhahir is with the Department of Electrical and Computer Engineering, University of Texas, Dallas, USA. (E-mail: aldhahir@utdallas.edu).}
}

\maketitle

\begin{abstract}
Semantic communication is envisioned as a promising technique to break through the Shannon limit. 
However, semantic inference and semantic error correction have not been well studied. 
Moreover, error correction methods of existing semantic communication 
frameworks are inexplicable and inflexible, which limits the achievable performance. In this paper, 
to tackle this issue, a knowledge graph 
is exploited to develop semantic communication systems. Two cognitive semantic 
communication frameworks are proposed for the single-user and multiple-user communication scenarios. 
Moreover, a simple, general, and interpretable semantic alignment algorithm for semantic information 
detection is proposed. Furthermore, an effective semantic correction algorithm is proposed by mining the 
inference rule from the knowledge graph. Additionally, the pre-trained model is fine-tuned to recover 
semantic information. For the multi-user cognitive semantic communication system, a message recovery 
algorithm is proposed to distinguish messages of different users by matching the knowledge level between 
the source and the destination. Extensive simulation results conducted on a public dataset demonstrate 
that our proposed single-user and multi-user cognitive semantic communication systems are superior to 
benchmark communication systems in terms of the data compression rate and communication reliability. 
Finally, we present realistic single-user and multi-user cognitive semantic communication systems results 
by building a software-defined radio prototype system.
\end{abstract}

\IEEEpeerreviewmaketitle
\section{Introduction}
\IEEEPARstart{W}{IRELESS} communication systems have been flourishing and evolved from the first 
generation (1G) to the latest fifth generation \cite{zhou2022cognitive}. Those systems have been 
developed based on the Shannon information theory \cite{wu2021unified}. With the development of advanced 
channel coding schemes, such as low-density parity-check (LDPC) codes and polar codes, traditional 
communication systems that transmit symbols have gradually approached the Shannon limit, which cannot meet 
future communication requirement of beyond 5G systems \cite{strinati20216g}. Moreover, due to the emerging 
diverse ultra-wide band communication services (such as online gaming, augmented reality (AR), virtual 
reality (VR), etc.), the increasing requirements of high data rate and the unprecedented proliferation of 
mobile devices, are confronted with the increasingly severe spectrum scarcity problem. Thus, it is 
imperative to make new breakthroughs and develop new communication  paradigms to improve spectrum 
efficiency \cite{shi2021new}.

Semantic communication proposed by Weaver and Shannon has been envisioned to tackle the above-mentioned 
challenges and has been receiving ever increasing attention from both academia and 
industry \cite{xu2011opportunistic}. Different from the conventional wireless communication systems that 
focus on the successful transmission of bits from the transmitter to the receiver, the semantic 
communication system pays more attention to the semantic information at the transmitter and the meaning 
interpreted at the receiver \cite{shi2021new}. It only transmits the indispensable information related to 
the specific task required at the receiver. Moreover, the  public and common human knowledge and 
experience as well as syntax, semantics, and inference rules can be exploited to further compress the 
transmitted information, improve the communication efficiency while guaranteeing the transmission 
reliability. Due to those advantages, recently, semantic communication has been increasingly 
investigated \cite{qin2021semantic}.

The investigations on semantic communication can be broadly categorized into three directions according to 
the role of the semantics, namely, semantics-oriented communication, goal-oriented communication 
and semantic-aware communication \cite{yang2022semantic}. In the semantic-oriented communication, 
how to efficiently extract semantic information at the transmitters and accurately recover it at the receiver are the focus.
Hence, the main challenges in the semantic-oriented communication systems lie in the data processing before 
sending and after receiving. Specifically, semantic encoding modules of the semantic-oriented communication systems are devised to obtain the significant semantic 
information of the data source and filter out unnecessary and redundant information. Meanwhile, semantic decoding
modules are exploited to accurately infer the semantic information. Different from the semantic-oriented
communication systems, the goal of the communication task in the goal-oriented communication systems 
plays an important role in semantic encoding and decoding. It can further filter out the goal-irrelevant
semantic information in each transmission. Essentially, goal-oriented communication is devised to accomplish
the task rather than the accurate recovery of the semantic information.
Basically, the core of both the semantic-oriented and goal-oriented communications is to establish a connection between 
two agents. In contrast, semantic-aware communication can be perceived to facilitate the completion of tasks.
In semantic-aware communication, the semantic information is obtained by analyzing the agent behavior and the current environment 
when performing the task instead of extracting it from a data source.

Recently, semantics-oriented communication has been increasingly investigated due to the advances in  
machine learning. It can be realized by two paradigms, namely, implicit reasoning and explicit reasoning 
 \cite{lu2022semantics}. Currently, deep learning (DL) technologies are leveraged extensively in 
most of the existing semantic communication systems to extract and reproduce the semantics of information
in an implicit reasoning process. Moreover, the joint optimization of the DL-based encoder and decoder
can achieve the end-to-end semantic delivery and recovery. Different from implicit reasoning, these schemes 
explicitly reason part of hidden semantics by introducing the reasoning mechanisms, such as KG-based 
inferring rules and probability theory-based semantic reasoning rules \cite{liang2022life}. The works related to those two 
paradigms are stated as follows.


For the first paradigm, the authors in \cite{farsad2018deep} proposed the joint source-channel coding
(JSCC) based on the bi-directional long short-term memory (BiLSTM). The text sentences were encoded into
fixed-length bits in simple channel environments. However, the fixed-length coding results in low
communication efficiency \cite{farsad2018deep}. In order to tackle this problem, the authors
in \cite{xie2021deep} proposed a powerful joint semantic-channel coding, namely DeepSC, to encode 
text information into various lengths under complex channel environments. It was shown that the 
utilization of various length coding can significantly improve the communication efficiency compared 
to the fixed-length coding in semantic communications. The semantic communication systems proposed 
in \cite{farsad2018deep} and \cite{xie2021deep} are appropriate for the receiver with a powerful 
computing capability that can implement a large-scale deep learning network and for the single-user 
communication scenario. Based on the works in \cite{farsad2018deep} and \cite{xie2021deep}, the 
authors in \cite{xie2020lite} proposed an environment-friendly semantic communication system, 
namely L-DeepSC, for the capacity-limited Internet-of-Things (IoT) devices while the authors 
in \cite{xie2021task} and \cite{xie2021task-O} proposed multi-user semantic communication systems, 
namely MU-DeepSC, for the multiple-user communication scenario transmitting single or multiple model data.

The second paradigm is still in the early research stage and is rarely studied since it requires
interdisciplinary knowledge. The authors in \cite{strinati20216g} presented a survey and proposed
that the knowledge graph can be used as the sharing knowledge system to realize semantic communication.
However, the specific framework and implementation were not presented. In \cite{shi2021new},
the concept of the semantic symbol was proposed. However, the specific implementation of the
semantic symbol abstraction and semantic symbol recognition were also not presented. To fill this gap,
we proposed a cognitive semantic communication framework in \cite{zhou2022cognitive}. We showed that
the exploitation of a knowledge graph can significantly improve the communication efficiency and
reliability. Our work in this paper is an extension of the work in \cite{zhou2022cognitive}. We
propose two cognitive semantic communication frameworks for the single-user and multiple-user
semantic communication scenarios. Furthermore, we propose an effective semantic correction algorithm
by mining the inference rules from the knowledge graph which allows the receiver to correct errors at
the semantic level.

Based on the comprehensive study of the above-mentioned related works, 
the advantages and disadvantages of those two paradigms are stated as follows. The first paradigm relies 
on the large-scale deep learning (DL) models to accurately and quickly recognize and extract intended 
semantic information. Nevertheless, DL normally requires a large number of high-quality labeled data. 
Moreover, the end-to-end operation is identified as a black-box process, which lacks interpretability. 
Even worse, the DL-based semantic encoder is only driven by data. Due to the deficiency of the guidance
from external knowledge, the DL-based semantic encoder is inflexible and redundant \cite{lu2022semantics}. 
The second paradigm achieves semantic compression and communication with the help of a knowledge base. 
Moreover, it has the interpretability during the compression process due to 
the reasoning rules in the knowledge base \cite{shi2021new}. It has the semantic information theory that 
paves the way for establishing theoretical analysis frameworks, which is of crucial importance in 
semantic communication \cite{seo2021semantics}. Unfortunately, to the best knowledge of the authors, 
few investigations have focused on this direction since it requires interdisciplinary knowledge, such 
as wireless communication, computer science, artificial intelligence, etc.

Motivated by the above-mentioned considerations, in this paper, two cognitive semantic communication systems
are proposed by making full use of the advantages of a knowledge graph. In addition, a knowledge graph-based explicit semantic reasoning technology
is leveraged to enhance semantic transmission by correcting the semantic errors using inference 
rules at the receiver in our proposed systems. In this sense, our proposed cognitive semantic communication
systems can be classified as semantics-oriented communication realized by the paradigm of explicit reasoning.
Notably, different from the traditional semantic communication system where the transmitter sends the abstraction of semantic information and
the receiver interprets its meaning by leveraging data-driven methods, our proposed cognitive semantic communication
systems have the \lq\lq cognitive\rq\rq \ characteristic, which is enabled by a knowledge graph. 
In the proposed systems, the semantic information is not required to be completely transmitted and only 
the important semantic information (such as the triplet, namely, the head entity, relation and tail entity) 
is transmitted. The receiver interprets the meaning by inferring from the received information. Due to 
the inference characteristic, our proposed cognitive semantic communication systems have three evident 
advantages. Firstly, a high semantic compression rate can be achieved since only the important information 
is transmitted. Secondly, a robust performance can be attained since the semantic error can be corrected 
via inference driven by the knowledge graph. Finally, the interpretability of the communication process 
can be realized since the triplets of the knowledge graph have the comprehensibility. The main contributions 
of this paper are summarized as follows.
\begin{itemize}
    \item To significantly improve the communication efficiency and reliability of the semantic
    communication system and tackle the problem that existing semantic communication systems based
    on the deep learning black-box framework lack interpretability, we exploit the knowledge graph
     to develop semantic communication systems. Moreover, a simple, general and
    interpretable algorithm for semantic information detection is proposed by exploiting triplets
    of the knowledge graph as semantic symbols. Furthermore, an effective semantic correction algorithm
    is proposed by mining the inference rules from the knowledge graph.
    \item Two cognitive semantic communication frameworks are proposed for the single-user and multi-user
    communication scenarios by exploiting the knowledge graph. Our proposed systems provide interpretable
    results since the triplets are general forms of semantic organization and are inherently readable.
    Moreover, the pre-trained model is fine-tuned in our two proposed frameworks to recover semantic
    information, which overcomes the drawback that a fixed bit length coding is used to encode sentences
    with different lengths. Furthermore, a message recovery algorithm is proposed for the multi-user
    cognitive semantic system to distinguish messages of different users by matching the knowledge level
    between the source and the destination.
    \item Extensive simulation experiments are performed on a public dataset. The simulation results
    demonstrate that our proposed single-user and multi-user cognitive semantic communication systems
    are superior to the benchmark communication systems in terms of the data compression rate and the
    communication reliability. Moreover, the simulation results also demonstrate the efficiency of our
    proposed semantic correction algorithm and message recovery algorithm. Furthermore, realistic
    single-user and multi-user cognitive semantic communication systems experiments are run on a
    software-defined radio (SDR) prototype that we built. The experimental results on the SDR demonstrate
    the feasibility of our proposed cognitive semantic communication systems.
\end{itemize}

The remainder of this paper is organized as follows. Section II presents the preliminary results and knowledge graph. By exploiting the knowledge graph, a single-user and a multi-user cognitive semantic communication system are proposed in Sections III and IV, respectively. Simulation results are presented in Section V. Finally, this paper is concluded in Section VI.
\section{Preliminary Results and Knowledge Graph}
\subsection{Semantic Representation Systems}
Different from traditional communication based on the Shannon's information theory that pursues the 
accurate transmission of bits, the goal of semantic communication is to achieve semantic equivalence 
between the source and destination. In semantic communication, semantic equivalence means that the 
meaning intended by the source is equivalent to the meaning understood by its destination \cite{strinati20216g}. 
It is not only influenced by the communication factors (such as channel fading, semantic coding, etc.), 
but also depends on the semantic representation and the knowledge level between the source and the destination. 
If the source and destination have the same or approximately same level of knowledge, a concise representation can be used by the source to achieve semantic equivalence. For example, the words \lq USA\rq \ and \lq America\rq \ have the same semantic information, but have completely different structures. \lq USA\rq \ instead of \lq America\rq \ can be used to represent and transmit the semantic information and achieve semantic equivalence if the source and destination have the same knowledge level about the \lq USA\rq \ and \lq America\rq. Thus, semantic representation is of crucial importance for the third paradigm of semantic communication, since it has an influence on the communication efficiency and semantic equivalence.

Semantic representation can be concise via a knowledge system since only the important semantic information is represented. In semantic communication, the semantic representation should satisfy three basic requirements. Firstly, the representation should reflect the logical relationship among semantic information. Secondly, the semantic representation should be concise. Finally, it does not perform complex  computational operations.

A knowledge graph is one of the most important semantic representation methods. It is a complete knowledge system that consists of computational ontology, facts, rules and constraints. Moreover, the inference rules for the knowledge graph have been well developed, such as  translational models \cite{bordes2013translating,wang2014knowledge,lin2015learning}, bilinear models \cite{trouillon2016complex} and deep learning models \cite{nguyen2017novel}, which can be beneficial for designing semantic correction algorithms and improving communication reliability. Thus, the knowledge graph is an important semantic representation method for semantic communication.
\subsection{Knowledge Graph}
Knowledge graph (KG), as the core of cognitive semantic communication, is a semantic network that reveals the relationship among entities in the form of graphs \cite{nguyen2017novel}. It generally consists of triplets (head entity, relation, and tail entity) \cite{wang2017knowledge}. A triplet is donated as $(h,r,t)$, where $h$, $r$ and $t$ denote the head entity, the relation and the tail entity of the knowledge graph, respectively. $h, t \in E$ and $r \in R$, where $E$ and $R$ are sets of entities and relations, respectively. Thus, a knowledge graph $KG$ can be denoted by ${\rm{KG  =  }}\langle {\rm E},{\rm R}\rangle$. Note that the set of entities include objects and concepts as well as their associated properties and values. The set of relations specify the relationship between entities. Generally speaking, relations have directions. The construction methods of a knowledge graph are usually divided into top-down and bottom-up methods \cite{bosselut2019comet}. A large-scale general knowledge graph used in our cognitive semantic communication is always constructed top-down including four procedures, namely, information extraction, knowledge fusion, knowledge processing, and knowledge update \cite{wu2014cognitive}. Since the focus of this paper is to investigate semantic communication, the details for knowledge graph construction are not given due to space limitation.
\section{Single-User Cognitive Semantic Communication System}
\begin{figure}[!t]
\centering
\includegraphics[width=4 in]{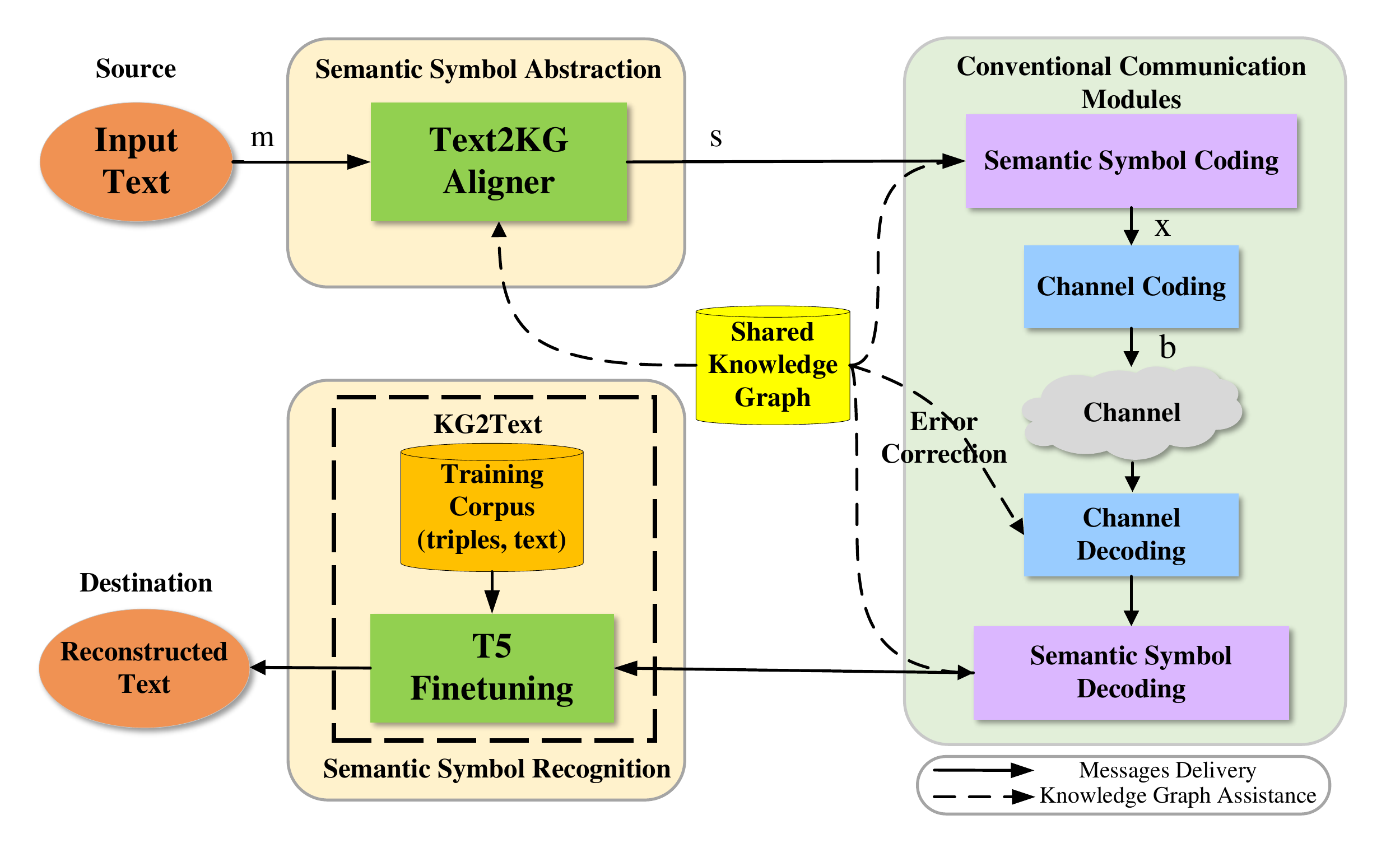}
\caption{The framework for our proposed single-user cognitive semantic communication system.} \label{fig.1}
\end{figure}
In this section, a single-user cognitive semantic communication system is proposed by exploiting the knowledge graph. The framework for this system and the details for the implementation are presented.
\subsection{Framework for The Single-User Cognitive Semantic Communication System}
The framework of our proposed single-user cognitive semantic communication system is shown in Fig. 1. A 
source $S$ generates a message $m\in M$. $M$ is the set of all the messages. $m$ carries the semantic 
information that $S$ transmits to $D$. In this paper, the text transmission is considered as an example. 
The semantic symbol is denoted as $s$, and $s$ is included in $S$, where $S$ is a set of the semantic 
symbols. In order to improve communication effectiveness, the semantic symbol is abstracted from the 
message $m$. This process is denoted as $s=Text2KG\left( m \right)$, where $Text2KG$ represents the Text2KG 
aligner. The Text2KG aligner is designed to extract the semantic symbol $s$ from 
message $m$. In this case, semantic compression is achieved.  Note that it should identify synonyms to 
align the same semantic words.

After the semantic symbol is obtained, the semantic symbol is transmitted by using the conventional 
communication modules (CCMs). Specifically, the semantic symbol $s$ is encoded as $x$ in order to improve 
the communication efficiency. Then, the channel coding is performed and $b$ is obtained. At the destination, 
the binary vectors are received and the semantic symbol code is obtained by channel decoding. Note that 
the semantic logic contained into the knowledge graph is exploited to correct errors in this process. In 
order to mine the semantic logic among the components of the triplets in the knowledge graph, triplets 
are embedded into the low-dimensional vector space by exploiting the representation learning of the 
knowledge graph. After that, the semantic symbol is obtained by using semantic symbol decoding.
\begin{figure}[htbp]
\centering
\includegraphics[width=2.5 in]{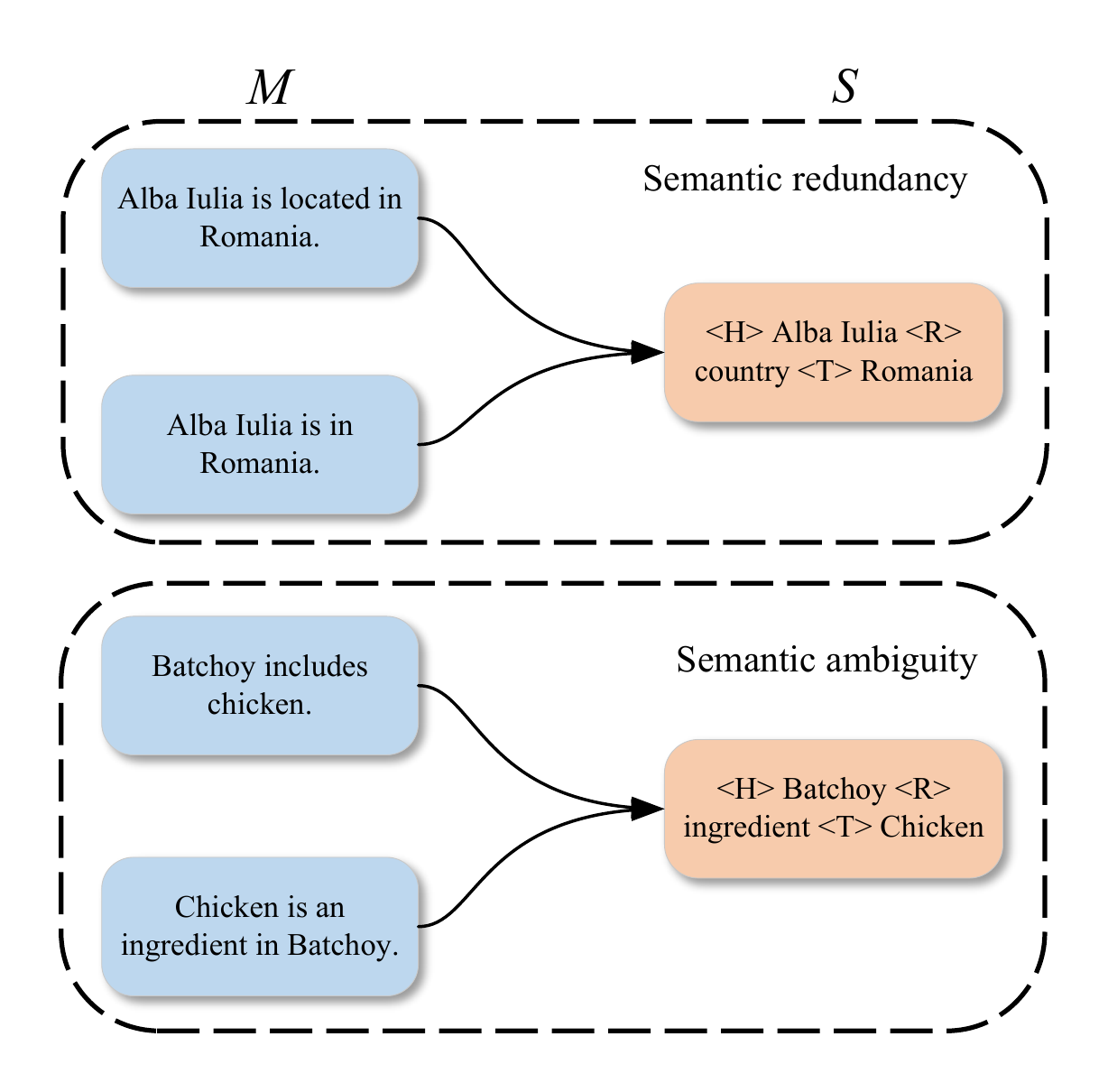}
\caption{An example shows the mapping from the message $m\in M$ to the semantic symbol $s\in S$.}
\label{fig.0}
\end{figure}
Finally, in order to reconstruct message $\hat{m}\in \hat{M}$ from the obtained semantic symbol, 
where $\hat{M}$ is the reconstructed message set, the semantic symbol recognition module is designed based 
on the natural language processing (NLP) technology. In this paper, different from the work in \cite{xie2021deep} 
that the joint source-channel codec (JSCC) framework is achieved by using the end-to-end training transformer, 
a novel framework by training a NLP technology at the receiver is proposed. In particular, depending on the length of the
sentence, a sentence could be mapped to one or more triplets by our proposed Text2KG aligner.  
Hence, compared with JSCC frameworks, our proposed framework can overcome the drawback that a fixed bit 
length coding is used to encode sentences with different lengths and results in low communication efficiency.

Note that since the mapping from $m$ to $s$ is many-to-one, the semantic ambiguity may exist. In order to 
mitigate semantic ambiguity and implement the triplet-to-text conversion, the pre-training model Text-to-Text 
Transfer Transformer model (T5) is fine-tuned on our training corpus. Since the pre-training model T5 is 
fed by billions of sentences, it can take context into account when the reconstructed text is generated. 
For example, as shown in Fig. 2, $S$ generates the message \lq Batchoy includes chicken\rq. \ However, \lq chicken\rq \ 
can be translated as animal or meat in different contexts. It results in the semantic ambiguity. The 
fine-tuned T5 model is used to reconstruct the text \lq Chicken is an ingredient of Batchoy.\rq \ at the 
receiver side. It is observed that the semantic ambiguity can be removed. In contrast, the conventional 
communication system cannot effectively tackle the semantic redundancy or semantic ambiguity.

\subsubsection{The Feasibility and Reasonability of Our Proposed Framework}
The semantic information theory is used to verity the feasibility and reasonability of our proposed cognitive semantic communication system. It is assumed that the source randomly transmits a message $m$ with the probability $\Pr(m)$. Thus, a massage entropy $H\left( M \right)$ of the source can be expressed as
$$H\left( M \right)=-\underset{m\epsilon M}{\mathop \sum }\,\Pr(m)\log_2\Pr(m).\eqno(1)$$
According to the work in \cite{shi2021new}, the logical probability $\Pr(s)$ of the semantic symbol $s$ can be expressed as
$$\Pr(s)=\underset{s=f\left( m \right),m\in M}{\mathop \sum }\,\Pr(m).\eqno(2)$$
Thus, based on Eq. (2), the semantic entropy $H\left( S \right)$ related to the semantic symbol set $S$ is
$$H\left( S \right)=-\underset{s\epsilon S}{\mathop \sum }\,\Pr(s)\log_2\Pr(s).\eqno(3)$$
According to the classic information theory, $H\left( S \right)$ can be written as
$$H\left( S \right)=H\left( S/M \right)+I\left( M;S \right),\eqno(4)$$
where $H\left( S/M \right)$ represents the uncertainty of the random variable $S$ under the condition of the known random variable $M$ and $I\left( M;S \right)$ represents the average mutual information between $S$ and $M$. Since $I\left( M;S \right)=H\left( M \right)-H\left( M/S \right)$, the semantic entropy can be rewritten as
$$H\left( S \right)=H\left( M \right)+H\left( S/M \right)-H\left( M/S \right).\eqno(5)$$
Since the mapping from $m$ to $s$ is often many-to-one by semantic symbol abstraction, $H\left( S \right)$ is always smaller than $H\left( M \right)$. Thus, there exists an entropy loss $H\left( M \right)-H\left( S \right)$. The reduced entropy of the source is a desired compression of the source without a \lq real\rq \ semantic loss, since the messages that contain the same semantics have many equivalent forms as shown in Fig. 2. For example, \lq Alba Iulia is located in Romania.\rq \ and \lq Alba Iulia is in Romania.\rq \ are different in the form. However, they have the same semantic information. Our proposed Text2KG aligner can align them to the same triplet \lq $<$H$>$ Alba Iulia $<$R$>$ country $<$T$>$ Romania\rq.
\subsection{The Implemention of Our Proposed Single-user Cognitive Semantic Communication System }
In this section, the details for the implementation of our proposed single-user cognitive semantic communication system are presented. An example for text transmission achieved by using our proposed single-user cognitive semantic communication system is shown in Fig. 3.
\begin{figure}[htbp]
\centering
\includegraphics[width=4 in]{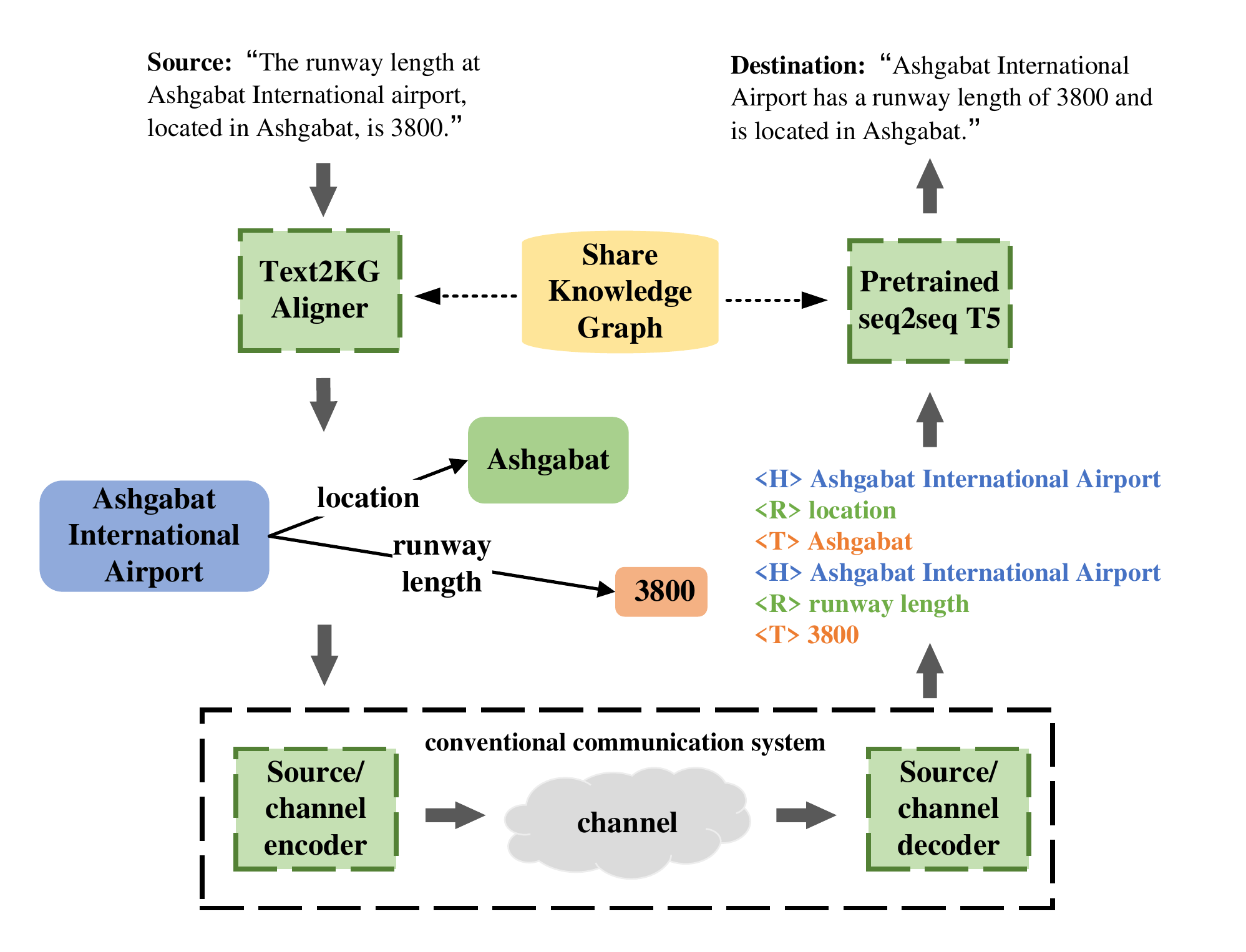}
\caption{An example for text transmission achieved by using our proposed single-user cognitive semantic communication system.}
\label{fig.2}
\end{figure}
\subsubsection{Semantic Symbol Abstraction}
The semantic symbol abstraction aims to detect and extract the specific semantic information contained in 
the message. The implementation of the semantic symbol abstraction is to align the input text $m$ with 
the triplet $(h,r,t)$. It is realized by a Text2KG alignment algorithm, as shown in Table 1. Note that 
for each sentence in the input text, all triplets that have $h$ and $t$ are matched. Moreover, each 
triplet can align to multiple sentences and each sentence can have multiple triplets aligned to it. The 
relations are not required to be matched since there are many ways to express them. Similar to the work 
in \cite{agarwal2020knowledge}, the relation can be expressed as long as the message mentions the head 
entity and tail entity.

Note that the alignment at the character level does not work in all cases.
For example, \lq the USA\rq \ and \lq America\rq \ have the same semantic meaning but are different words.
 They cannot be aligned at the character level. In order to overcome this drawback, a synset is exploited
 to query all synonyms of an entity. In this paper, the WordNet is exploited as a synset. It is a large
 database of English words built in the 1980s by George Miller's team at Princeton University
 \cite{miller1995wordnet}. Nouns, verbs, adjectives and adverbs are stored in this database.
 All the synonyms for head entities and tail entities are taken into account by our Text2KG alignment
 algorithm. Thus, the input text is aligned with the triplets at the semantic level.
\begin{table}[htbp]
\begin{center}
\caption{The semantic alignment algorithm}
\begin{algorithm}[H]
\SetAlgoLined
\caption{The alignment algorithm}
\KwIn{s is the sentences of the input text,\\
\ \ \ \ \ \ \ \ \ (h, r, t) is the triplets of the knowledge graph.}
\SetKwFunction{Fs}{synonym}
\SetKwProg{Fn}{Function}{:}{}
\Fn{\Fs{word}}{
var wordset:list\\
wordset initalize\\
\tcp{word.synsets means the synonyms set consisting of all synonyms for this word.}
\eIf{word.synsets is not empty}{
wordset.add(word.synsets)}{
wordset = word}
}
\KwResult{wordset}
var alignment triplets:list\\
alignment triplets initalize\\
\For{all s $\in$ input text}{
\For{all triplets (h, r, t) $\in$ KG}{
\If{s.contains(synonym(h)) and s.contains(synonym(t))}{
alignment triplets.add((h, r, t))}}
}
\KwOut{alignment triplets}
\end{algorithm}
\end{center}
\end{table}
\subsubsection{Conventional Communication Modules}
After the semantic symbols are obtained, in order to complete semantic communications, those symbols are 
transmitted by using the conventional communication modules, namely, semantic symbol coding, channel 
coding, transmission channel, channel decoding and semantic symbols decoding. Specifically, according 
to \cite{zhou2021resource}, for a propositional logic with finite propositions, the size of all possible 
interpretations (worlds) is finite. Thus, the number of entities and relations in the knowledge graph 
which describes the world is finite and it ensures that the semantic symbol code has a limited length. 
Firstly, in order to achieve semantic symbol coding, a dictionary $\left( key, value \right)$ is created. 
It enables the head entity, tail entity and relation of the knowledge graph to be uniquely mapped as 
integers. Secondly, a binary vector $x$ is obtained by encoding each integer with a fixed length. 
This process is denoted as the semantic symbol encoding (SSC).

\subsubsection{Semantic Correction Modules}
Moreover, in order to improve the communication reliability, the channel coding module is applied
and the coded binary vector set $B$ is obtained. Then, the set $B$ is transmitted over the channel.
Note that the binary channel is considered in this paper. Since the channel is noisy,
transmission errors can occur during transmission. In order to improve the robustness of our proposed
system, channel decoding is implemented by exploiting the knowledge graph to correct transmission errors.
Our correction algorithm is based on inference rules obtained by mining the knowledge graph.
They can evaluate the semantic reasonableness of a triplet consisting of $h \in E$, $r \in R$, $t \in E$,
or obtain the most relevant hidden entities and relations of a given entity recognized from the source
signal \cite{bordes2013translating}.

Let $\Phi _t$ be the set of all triplets in the message at the time slot $t$. The $i$th triplet at the 
time slot $t$ is denoted as $\phi _t^i = \langle e_{i,t}^h,{r_{i,t}},e_{i,t}^t\rangle$
where $\phi _t^i \in {\Phi _t}$, $e_{i,t}^h, e_{i,t}^t \in E$, $r_{i,t} \in R$. $e_{i,t}^h$, $r_{i,t}$ 
and  $e_{i,t}^t$ denote the head entity, relation and the tail entity of the $i$th triplet at the time 
slot $t$. The key of our correction algorithm is to mine the relationships among the observable and 
hidden entities as well as those among entities and relations. It can automatically judge the rationality 
of triplets at the destination and correct the triplets that are inconsistent with the semantic logic. 
For example, the message \lq I eat apple\rq \ is transmitted at the source while the message \lq I eat 
knife\rq \ is received at the destination. In this case, our correction algorithm is active, since \lq I 
eat knife\rq \ is inconsistent with the semantic logic.

In our correction algorithm, the representation learning of knowledge graph is exploited to mine the 
relationships among the entities and relations. Particularly, an embedding-based method is propoesd to evaluate 
rationality semantically of the $(h, r, t)$ at the destination. The main idea of this method is to learn 
a mapping function that maps the entities and relations in the high-dimensional graphical representation 
space into a low-dimensional embedding space. The low-dimensional representation of the entities and 
relations, also called entity embedding and relation embedding, retains the information about the structure 
of the graph and is easy to be used for inference. The embedding of entities $e_t$ and relations $r_t$ are 
denoted by $\vec{e_t}$ and $\vec{r_t}$, respectively, and $\vec{\phi _t^i}$ denotes the triplet of embedding. 
The inference function, denoted as $f(\vec{\phi _t^i})$, is introduced to measure the rationality of the 
triplet. Our obtained inference function has the following two characteristics.
\begin{enumerate}
    \item If $\phi _{\rm{i}}^t = \langle e_t^s,{r_t},e_t^o\rangle$ is consistent with the inference logic contained in the knowledge graph, the score of $f(\vec{\phi _t^i})$ is maximized.
    \item If $\widetilde {\phi _{\rm{i}}^t} = \langle \widetilde {e_t^s},\widetilde {{r_t}},\widetilde {e_t^o}\rangle$ is inconsistent with the inference logic of the knowledge graph, the score of $
        f(\vec{\widetilde {\phi _t^i}})$ is much smaller than $f(\vec{\phi _t^i})$. Moreover, the difference between $f(\vec{\widetilde {\phi _t^i}})$ and $f(\vec{\phi _t^i})$ is proportional to their meaning difference.
\end{enumerate}

The principle of the correction is to traverse the semantic symbols in the knowledge graph in order to find the most similar and reasonable semantic symbol observed at the receiver. The details for our proposed correction algorithm denoted as Algorithm 2 are given in Table II. Semantic symbol decoding is the reverse process of semantic symbol encoding. After semantic symbol decoding, the reconstructed semantic symbol $\hat{S}$ is achieved.
\begin{table}[htbp]
\begin{center}
\caption{The semantic correction algorithm}
\begin{algorithm}[H]
    \SetAlgoLined
    \caption{The semantic correction algorithm}
    \KwIn{$\hat{h}$ is the binary vector of the head entity observed at the receiver, \\
          \ \ \ \ \ \ \ \ \ $\hat{r}$ is the binary vector of the relation observed at the receiver, \\
          \ \ \ \ \ \ \ \ \ $\hat{t}$ is the binary vector of the tail entity observed at the receiver. \\ 
          \ \ \ \ \ \ \ \ \ H is the head entities set of the knowledge graph. h $\in$ H is the head entity of a triplet . $h$ is the channel coding of h. \\
          \ \ \ \ \ \ \ \ \ R is the relations set of the knowledge graph. r $\in$ R is the relation of a triplet. $r$ is the channel coding of r. \\
          \ \ \ \ \ \ \ \ \ T is the tail entities set of the knowledge graph. t $\in$ T is the tail entity of a triplet. $t$ is the channel coding of t. \\}
    \SetKwFunction{Fc}{candidate}
    \SetKwProg{Fn}{Function}{:}{}
    \Fn{\Fc{hs, rs, ts}}{
        var triplets, temp:list\\
        triplets initialize\\
        \For{head $in$ hs}{
        temp initialize\\
        temp.add(head)\\
        \For{rel $in$ rs}{
        temp=temp[:1]\\
        temp.add(rel)\\
        \For{tail $in$ ts}{
        temp.add(tail)\\
        triplets.add(temp)\\
        temp=temp[:-1]
        }}
        }
    }
    \KwResult{triplets}
    \SetKwFunction{Fs}{SimMax3}
    \SetKwProg{Fn}{Function}{:}{}
    \Fn{\Fs{$\hat{h}$, $\hat{r}$, $\hat{t}$}}{
    var Hsim, Rsim, Tsim, M3H, M3R, M3T:float\\
    Hsim, Rsim, Tsim = similar($\hat{h}$, $h$), similar($\hat{r}$, $r$), similar($\hat{t}$, $t$) \\ \tcp{The cosine similarity of two vectors is achieved by Function similar.}
    M3H, M3R, M3T = max3(Hsim), max3(Rsim), max3(Tsim) \\ \tcp{The top 3 element are found by Function max3.}
    hs, rs, ts = sim2tri(M3H), sim2tri(M3R), sim2tri(M3T) \\ \tcp{The entities and relations corresponding to the similarity are found by Function sim2tir.}
    }
    \KwResult{hs, rs, ts}
    hs, rs, ts = SimMax3($\hat{h}$, $\hat{r}$, $\hat{t}$)\\
    $\Phi$ = candidate(hs, rs, ts)\\
    score = $f(\phi )$, $\phi  \in \Phi$ \\
    TriCor = $\Phi$[score.index(max(score))]\\
    \KwOut{TriCor}
\end{algorithm}
\end{center}
\end{table}
\subsubsection{Semantic Symbol Recognition}
After the semantic symbol is obtained, in order to interpret the meaning of semantic information, the 
crucial process of our proposed system is to transform triplets into text. This process needs to exploit 
the NLP technology. In this paper, the T5 model \cite{raffel2019exploring} is fine-tuned on the local 
dataset in order to convert triplets in the knowledge graph into natural text. Specifically, the triplets 
in the dataset need to be concatenated as $head \ relation\_1 \ tail\_1,....,relation\_n \ tail\_n$ for 
inputing into T5.
\subsubsection{Model Training}
Our proposed cognitive semantic communication system is not an end-to-end training system. It includes 
two large scale deep neural networks, namely, an embedding model and the T5 model at the destination. 
The details for our proposed single-user system training process are given in Table III. The training 
process of our proposed cognitive semantic communication consists of two phases. The first phase is to 
train the embedding model and the second phase is to train the T5 model.
\begin{table}[htbp]
\begin{center}
\caption{Single-user system training process}
\begin{algorithm}[H]
\SetAlgoLined
\caption{Single-user system training process}
\textbf{Initialization:} The training dataset $K{G^T}$ and the model hyperparameter: batch size $B$, epoch $n$, learning rate $lr$.\\
\SetKwFunction{Fte}{Train Embedding Model}
\SetKwProg{Fn}{Function}{:}{}
\Fn{\Fte{}}{
    \textbf{Input:} Choose mini-batch data $\left\{ \Phi  \right\}_{{\rm{j}} = n}^{n + B}$ from $K{G^T}$. \\
    $\widetilde \Phi$ = NegativeSampler($\Phi$) \\
    \For{j = n $\to$ n + $B$}{
        $Mode{l_{emb}}$($\phi$, $\widetilde \phi$) $\to$ pos, neg \\
        Compute $\rm{loss}$ by pos and neg. \\
        Optimize $Mode{l_{emb}}$ $\to$ Gradient descent with $\rm{loss}$ by $lr$\\
    }
    \KwResult{$Mode{l_{emb}}$}
}
\textbf{Initialization:} The training dataset ${D^T}$ and the model hyperparameter: batch size $B$, epoch $n$, learning rate $lr$ \\
Download the pre-training T5 model \\
\SetKwFunction{Ft}{Finetune T5 Model}
\SetKwProg{Fn}{Function}{:}{}
\Fn{\Ft{}}{
\textbf{Input:} Choose mini-batch $\left\{ {{\rm{(}}{{\rm{t}}^T},{s^T})} \right\}_{j = n}^{n + B}$ from ${D^T}$ \\
\For{j = n $\to$ n + B}{
    $Mode{l_{T5}}$$({t^T},{s^T})$ $\to$ $loss$\\
    Optimize $Mode{l_{T5}}$ $\to$ Gradient descent with $loss$ by $lr$\\
 }
\KwResult{$Mode{l_{T5}}$}
}
\SetKwFunction{Fw}{The whole Network}
\SetKwProg{Fn}{Function}{:}{}
\Fn{\Fw{}}{
\textbf{Input text}: $m$ $\in$ $M$ \\
\textbf{Semantic symbols}: $s$ = Text2KG(m) \\
Transmit $s$ by conventional communication modules and achieve $\tilde{s}$ at the destination. \\
\textbf{Corrected Semantic symbols}: $\hat{s}$ = correction($\tilde{s}$, $Mode{l_{emb}}$) \\
\textbf{Output text}: output = KG2Text($\hat{s}$, $Mode{l_{T5}}$)
}
\end{algorithm}
\end{center}
\end{table}

Similar to the most existing training methods of the embedding model, two non-overlappering training 
sample sets, namely, a set of positive samples and a set of negative samples, are constructed. The positive 
samples are the real triplets of the knowledge graph. The negative samples are constructed by replacing 
the head entity or tail entity of the real triplets with the wrong one. The scoring functions are defined 
as the loss functions of the embedding model and the training objective is to minimise the scoring functions.
Specifically, different embedding models have different scoring functions \cite{ali2021bringing}. The standard 
stochastic gradient descent approach is exploited to train the embedding model. According to the work in \cite{trouillon2016complex}, 
the above process finally converges to a stationary solution with the relative distances among the embedding values of different 
entities and relations. In addition, the real triplets obtain higher scores than the wrong triplets by scoring 
functions after training the embedding model. Correspondingly, the real triplets are consistent with the 
semantic logic, and the wrong triplets are inconsistent with the semantic logic. In this sense, 
the gap between the scores of the triplets reflects distance in semantic difference. Using the standard 
stochastic gradient descent approach, T5 model is fine-tuned in order to realize text recovery.
\begin{figure}[htbp]
\centering
\includegraphics[width=4 in]{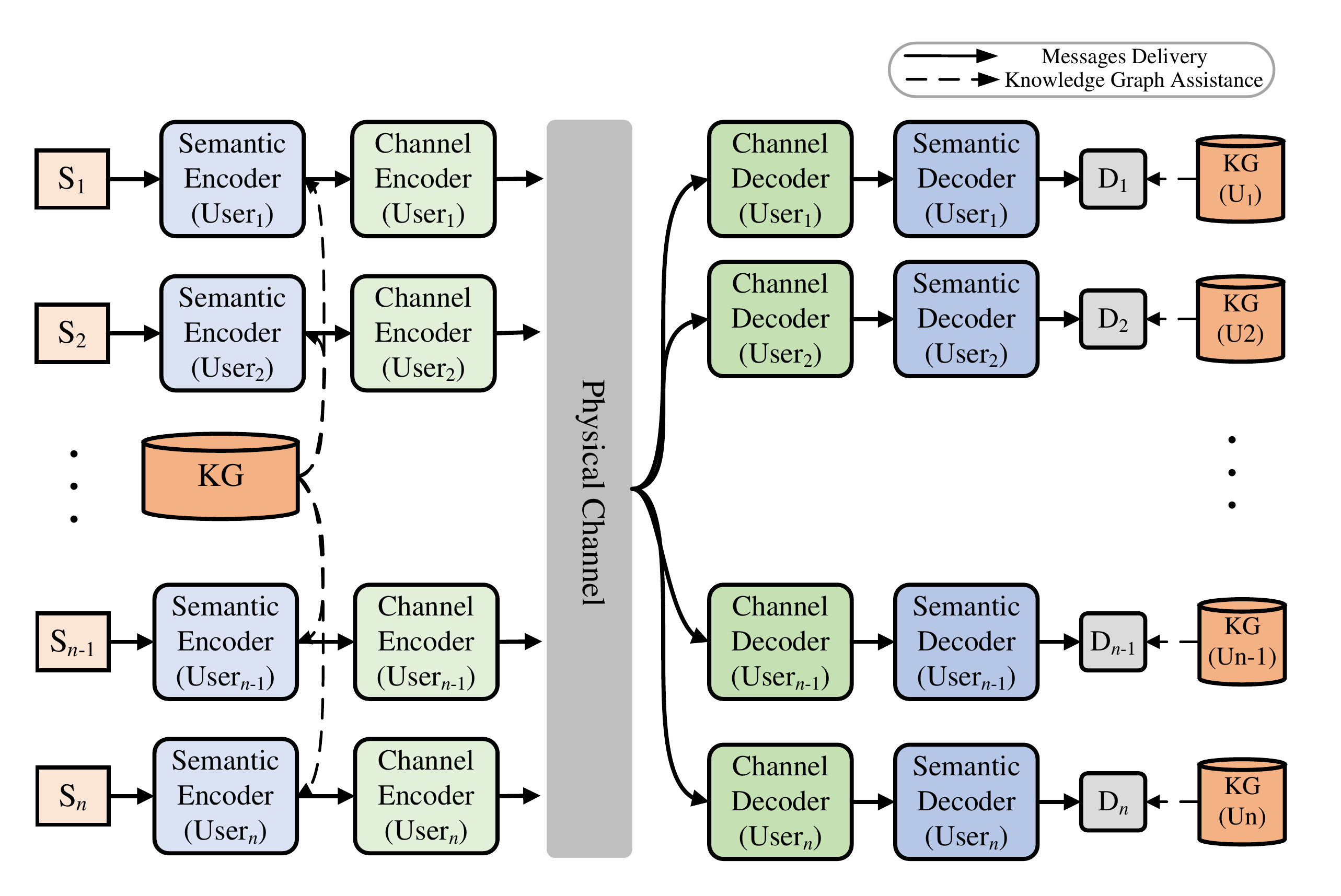}
\caption{The framework for our proposed multi-user cognitive semantic communication system.}
\label{fig.2}
\end{figure}
\section{Multi-User Cognitive Semantic Communication System}
Based on the single-user cognitive semantic communication system, considering the fact that the multiple-user semantic communication can further improve the communication efficiency, in this section, a multi-user cognitive semantic system is proposed by using the knowledge graph. The details for its framework and implementation are presented. Moreover, a message recovery algorithm is proposed in order to distinguish the messages of different users. The framework for this system is shown in Fig. 4.
\subsection{Framework for The Multi-user Cognitive Semantic Communication System}
As shown in Fig. 4, a multi-user cognitive semantic communication system that consists of $n$ sources and $n$ destinations is considered. The message of the $k$th source is denoted as
${m_k}$. Each source transmits semantic information. Similar to the single-user cognitive semantic communication system, the semantic information is first expressed as the semantic symbols, namely triplets. The semantic symbol ${s_k}$ is abstracted from the message ${m_k}$ by using our proposed Text2KG aligner, given as
$$ {s_k} = Text2KG({m_k}).\eqno(6)$$

After the semantic symbols of each source are obtained, they are transmitted by exploiting the conventional communication modules (CCMs). Specifically, the semantic symbol $s_k$ is encoded in order to improve the transmission efficiency. Then, the channel coding is performed and $b_k$ is obtained.
$$ {b_k} = C({s_k}),k = 1,2,...,n\eqno(7)$$
where $C$ is the channel coding; $b_k$ and $s_k$ are the channel coding and semantic symbols of the $k$th source, respectively.

After transmission over the channel, the channel decoding is performed at each user receiver. The reconstructed semantic symbol $\widehat {{s_k}}$ is obtained by exploiting our proposed correction algorithm. Note that the semantic symbols of different users are not distinguished in this process. Thus, the reconstructed semantic symbol of each user is mixed, given as
$$\widehat {{s_k}} = correct({C^{ - 1}}(\widehat {{b_k}})),k = 1,2,...,n\eqno(8)$$
where $C^{ - 1}$ is the channel decoding; $correct$ is our proposed correction; $\widehat {{s_k}}$ is the reconstructed semantic symbol and $\widehat {{b_k}}$ is the channel coding received at the $k$th user.

Then, the reconstructed semantic symbols of each user need to be distinguished. Since different users have their own knowledge graphs that are different from other users' knowledge graphs in practice, each user can only understand semantic information which match their own knowledge graph \cite{bao2011towards}. Thus, the private knowledge graph of the $k$th user is exploited to distinguish the private restructured semantic symbol $\widehat{s_k}$. Finally, similar to the single-user cognitive semantic communication system, the restructured message $\widehat{m_k}$ is obtained by exploiting our proposed fine-tuned T5 model.
\subsection{The Implemention of the Multi-User Cognitive Semantic Communication System}
In this section, the details for the implementation of our proposed multi-user cognitive semantic
communication system are presented. An example for text transmission achieved by using our proposed
multi-user cognitive semantic communication system is shown in Fig. 5.
\begin{figure}[htbp]
\centering
\includegraphics[width=4 in]{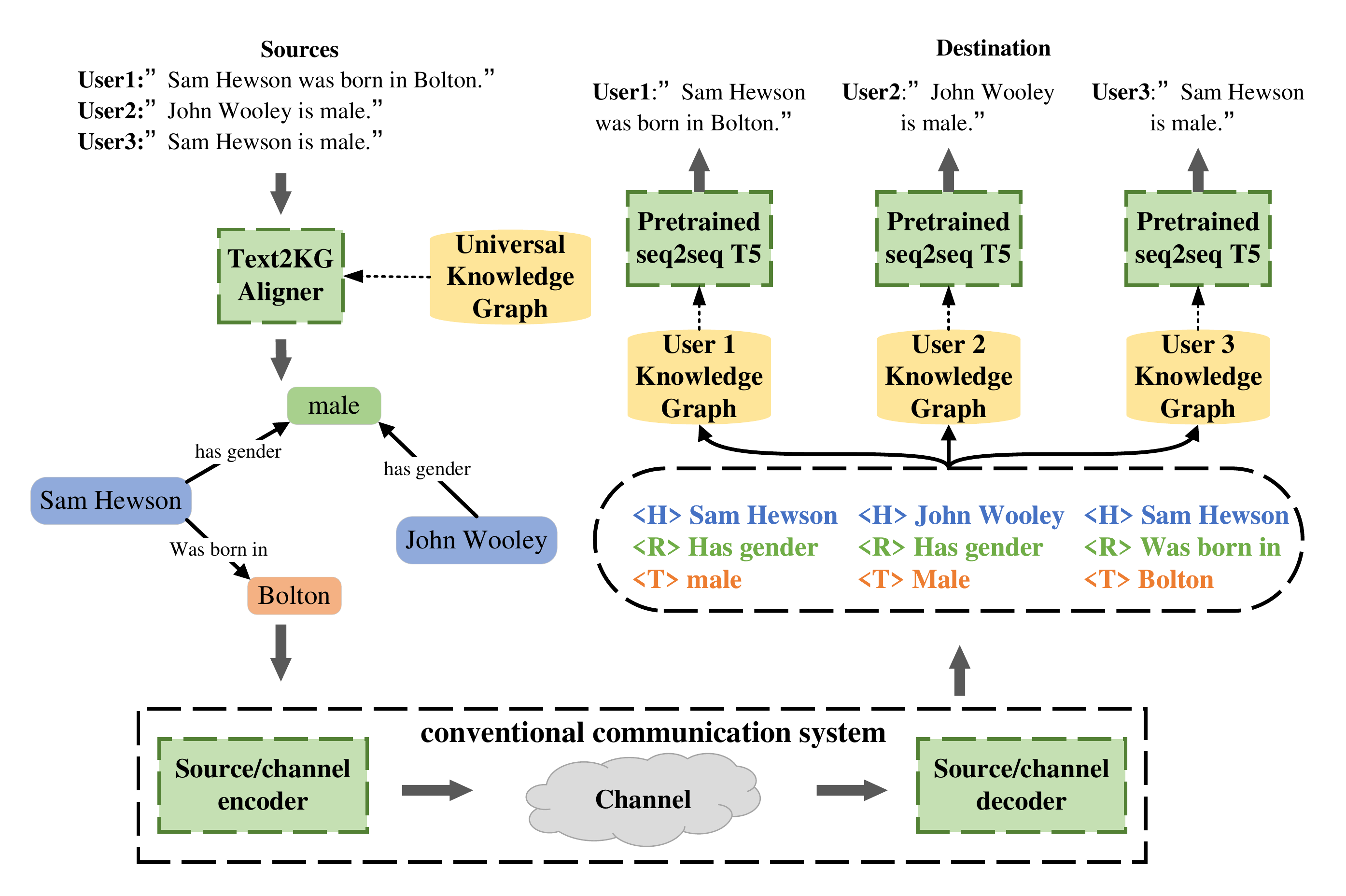}
\caption{An example for text transmission achieved by using our proposed multi-user cognitive semantic communication system.}
\label{fig.6}
\end{figure}
\subsubsection{Multi-user Cognitive Semantic Transmitter}
Different sources generate different messages. Similar to the single-user cognitive semantic communication system, the semantic symbol of each user is abstracted from the message by exploiting our proposed Text2KG alignment algorithm. Then, those symbols are transmitted by using the conventional communication modules.
\subsubsection{Multi-user Cognitive Semantic Receiver}
At the destination, the restructured semantic symbols are achieved by exploiting our proposed correction algorithm. Since the semantic symbols of all users are mixed together, a message recovery algorithm is proposed to distinguish the semantic symbols based on the private knowledge graph. Specifically, for the $k$th user, all the restructured semantic symbols are traversed to judge which belongs to the $k$th user's private knowledge graph. The details for the message recovery algorithm denoted as Algorithm 3 can be found in Table IV. Similar to the single-user cognitive semantic communication system, the fine-tuned T5 model is exploited to convert the restructured semantic symbols into the restructured messages.
\begin{table}[htbp]
\begin{center}
\caption{The message recovery algorithm}
\begin{algorithm}[H]
    \SetAlgoLined
    \caption{The message recovery algorithm}
    \KwIn{$\widehat s$ is the restructured semantic symbols at the destination.\\
    \ \ \ \ \ \ \ \ $KG\_{u_k}$ is the $k$-th user's private knowledge graph.}
    \SetKwFunction{Fd}{distinction}
    \SetKwProg{Fn}{Function}{:}{}
    \Fn{\Fd{$\widehat s$, $KG\_{u_k}$}}{
        SimMax initialize\\
        \For{$s$ $in$ $\widehat s$}{
        sim initialize\\
        \For{$t$ $in$ $KG\_{u_k}$}{
        sim.add(similar($s$, $t$))}
        SimMax.add(max(sim))
        }
    }
    \KwResult{$\widehat s$ [SimMax.index(max(SimMax))]}
    \KwOut{distinction($\widehat s$, $KG\_{u_k}$)}
\end{algorithm}
\end{center}
\end{table}
\section{Simulation Results}
In this section, simulation results are presented to evaluate the performance of our proposed single-user 
and multi-user cognitive semantic communication systems. They are compared with DNN-based methods and the traditional 
communication systems realized by the separate source and channel coding technologies. The simulation experiments are performed under the binary 
symmetric channels (BSC), additive gaussian white noise (AWGN) channels and Rayleigh fading channels. Moreover, realistic single-user and multi-user cognitive semantic communication 
systems experiments are run on our developed SDR prototype systems.
\subsection{Dataset and Simulation Setup}
Yet Another Great Ontology (YAGO) \cite{rebele2016yago} is a knowledge graph containing facts that are 
extracted from  Wikipedia and aligned with WordNet in order to exploit the large amount of information 
included in WordNet. It contains general facts about public figures, geographical entities, movies, and 
further entities, and has a taxonomy for those concepts. YAGO3-10 is a subset of YAGO3 \cite{mahdisoltani2014yago3} 
(an extension of YAGO) that contains entities associated with at least ten different relations. In total, 
YAGO3-10 has 123182 entities and 37 relations, and most of the triplets describe attributes of persons 
such as citizenship, gender, and profession. The number of triplets in the training and those in the test 
sets of the YAGO3-10 are presented in the Table V. In this paper, ComplEx model is selected as the embedding 
model, since it is one of the top-ten-performing interaction models on YAGO3-10 \cite{ali2021bringing}. 
The ComplEx model is trained on the training set for 1000 steps with a learning rate of 0.0004 and the 
embedding dimension is 100.
\begin{table}[htbp]
\begin{center}
\caption{Number of triplets in the training and the test sets of the YAGO3-10.}
\begin{tabular}{lcl}
\\\toprule
$\textbf{Split}$\ \ \ \ \ \ \ \ \ \ \ \ \ \ \ \ \ \ \ \ \ \ \ \ \ \ \ \ \ \ \ \ \ \ \ \ \ \ \ \ \ \ \ $\textbf{Number of Samples}$\\ \midrule
Train(Triplets)\ \ \ \ \ \ \ \ \ \ \ \ \ \ \ \ \ \ \ \ \ \ \ \ \ \ \ \ \ \ \ \ \ \ \ \ \ \ \ 1079029\\
Test(Triplets)\ \ \ \ \ \ \ \ \ \ \ \ \ \ \ \ \ \ \ \ \ \ \ \ \ \ \ \ \ \ \ \ \ \ \ \ \ \ \ \ \ \ 2000\\
\bottomrule
\end{tabular}
\end{center}
\end{table}
The training set of WebNLG English dataset \cite{agarwal2020knowledge} is used to fine-tune the T5 model. It contains knowledge graphs and text from various domains including Airport, Artist, Astronaut, Athlete, Building, Celestial Body, City, Comics Character, Food, Mode of Transportation, Monument, Politician, Sports Team, University and Written Work \cite{agarwal2020knowledge}. The test sets include three additional domains, namely, Film, Scientist, and Musical Work. The number of texts and knowledge graph pairs in the training, test sets of the WebNLG English dataset is presented in Table VI. The T5 model is fine-tuned on the training sets for 100 steps with a learning rate of 0.0001. We compare the performance of the proposed system with the benchmark systems by transmitting the text in the test set.
\begin{table}[htbp]
\begin{center}
\caption{Number of texts and knowledge graph pairs in the training and the test sets of the WebNLG English dataset.}
\begin{tabular}{lcl}
\\\toprule
$\textbf{Split}$\ \ \ \ \ \ \ \ \ \ \ \ \ \ \ \ \ \ \ \ \ \ \ \ \ \ \ \ \ \ \ \ \ \ \ \ \ \ \ \ \ \ \ $\textbf{Number of Samples}$\\ \midrule
Train(T-G Pairs)\ \ \ \ \ \ \ \ \ \ \ \ \ \ \ \ \ \ \ \ \ \ \ \ \ \ \ \ \ \ \ \ \ \ \ \ \ \ \ 35426\\
Test(T-G Pairs)\ \ \ \ \ \ \ \ \ \ \ \ \ \ \ \ \ \ \ \ \ \ \ \ \ \ \ \ \ \ \ \ \ \ \ \ \ \ \ \ \ 4464\\
\bottomrule
\end{tabular}
\end{center}
\end{table}
\subsection{Performance Metric}
In traditional communication systems, the symbol error rate (SER) is used to evaluate the communication performance. However, it is inappropriate for semantic communication since different sentences at the transmitter and receiver possibly have the same semantic information \cite{xie2021deep}. In order to tackle this issue, the Bilingual evaluation understudy (BLEU) and semantic similarity score are widely used to evaluate the performance of semantic communication and that of the traditional communication. BLEU score is usually used to evaluate the machine translation results. However, the BLEU score can only compare the difference among words of two sentences rather than their semantic information. Thus, in this paper, the semantic similarity score is used to measure performance \cite{xie2021deep}. According to the cosine similarity, the sentence similarity between the original sentence $m$ and the recovered sentence $\hat{m}$ is calculated as:
$$score\left( \hat{m},m \right)=\frac{{{B}_{\text{ }\!\!\Phi\!\!\text{ }}}\left( m \right){{B}_{\text{ }\!\!\Phi\!\!\text{ }}}{{\left( {\hat{m}} \right)}^{T}}}{\left| \left| {{B}_{\text{ }\!\!\Phi\!\!\text{ }}}\left( m \right)\left| \left| ~ \right| \right|{{B}_{\text{ }\!\!\Phi\!\!\text{ }}}\left( {\hat{m}} \right) \right| \right|},\eqno(9)$$
where ${{B}_{\text{ }\!\!\Phi\!\!\text{ }}}$, representing bidirectional encoder representations from transformers (BERT) is a large pre-trained model used for extracting the semantic information. It is trained on an ultra large-scale corpus. Compared with the BLEU score, BERT has been fed by billions of sentences \cite{xie2021deep}. Therefore, it can recognize semantic information effectively.
\subsection{Performance Comparison}
Without loss of generality, our proposed cognitive semantic communication systems are compared with DNN-based approaches and the traditional separate source coding and channel coding
approaches under the BSC channels, AWGN channels and Rayleigh fading channels. Specifically, the DNN-based approaches, such as the joint source-channel 
coding (JSCC) approach \cite{8461983} and the DeepSC approach \cite{xie2021deep}, are compared with
our proposed single-user cognitive semantic communication system. Besides, our proposed multi-user cognitive 
semantic communication system is compared with the MR-DeepSC approach \cite{hu2022one}. Our proposed systems and other DNN-based benchmark systems are presented in Table VII.
To guarantee the fairness of performance comparison, the channel coding scheme of our proposed 
system is the same as that of the benchmark traditional communication systems, namely, the binary 
convolutional codes (BCC). Moreover, two source coding schemes are considered for the benchmark 
systems, namely, Huffman and fixed-length coding. In this paper, the fixed-length coding scheme adopts 
7 bits for each character since the corpus is case-sensitive and contains special characters. It is well 
known that the variable-length coding is more efficient than the fixed-length coding. To demonstrate the 
superiority, the fixed-length coding is chosen as the source coding of our proposed system.
\\
\begin{table}
    \normalsize\centering
    \begin{center}
    \renewcommand\arraystretch{1}
    \setlength{\tabcolsep}{0.7mm}
    \newcommand{\tabincell}[2]{\begin{tabular}{@{}#1@{}}#2\end{tabular}}
    \resizebox{\textwidth}{!}{\begin{tabular}{c|c|c}
    \Xhline{2pt}
    Single-user Models & General Features & Technical Methods \\
    \Xhline{1pt}
    \tabincell{c}{Single-user Cognitive Semantic \\ Communication System} & \tabincell{c}{knowledge graph\\ semantic alignment algorithm \\ semantic correction algorithm} & \tabincell{c}{(1) Our proposed Text2KG alignment algorithm is \\ exploited to achieve semantic extraction.\\ (2) Our proposed semantic correction algorithm is exploited to \\ correct semantic errors and recover semantic information.}\\
    \Xhline{1pt}
    DeepSC \cite{xie2021deep} & \tabincell{c}{ the Transformer model\\ transfer learning \\ mutual information } & \tabincell{c}{(1) The encoder and decoder of Transformer are \\ exploited to achieve semantic encoding and decoding. \\ (2) Three dense layers are exploited to \\estimate the mutual information.} \\
    \Xhline{1pt}
    JSCC \cite{8461983} &\tabincell{c}{joint source-channel coding \\ the LSTM model.}& \tabincell{c}{(1) BiLSTM model is leveraged to achieve semantic encoder. \\ (2) LSTM model is leveraged to achieve semantic decoder.}\\
    \hline\hline
    Multi-user Models & General Features & Technical Methods \\
    \Xhline{1pt}
    \tabincell{c}{Multi-user Cognitive Semantic\\ Communication System} & \tabincell{c}{knowledge graph\\ message recovery algorithm} &\tabincell{c}{A message recovery algorithm is proposed to distinguish \\messages of different users by matching the knowledge level.}\\
    \Xhline{1pt}
    MR-DeepSC \cite{hu2022one} & \tabincell{c}{DistilBERT \\ semantic recognizer} & \tabincell{c}{By leveraging semantic features for different users, a \\ semantic recognizer based on the DistilBERT is built.}\\
    \Xhline{2pt}
    \end{tabular}
}
\caption{Our proposed models and other DNN-based benchmark systems.}
\end{center}
\end{table}

\begin{figure}
\centering
\subfigure[] {
    \label{fig:a}
    \includegraphics[width=2.5 in]{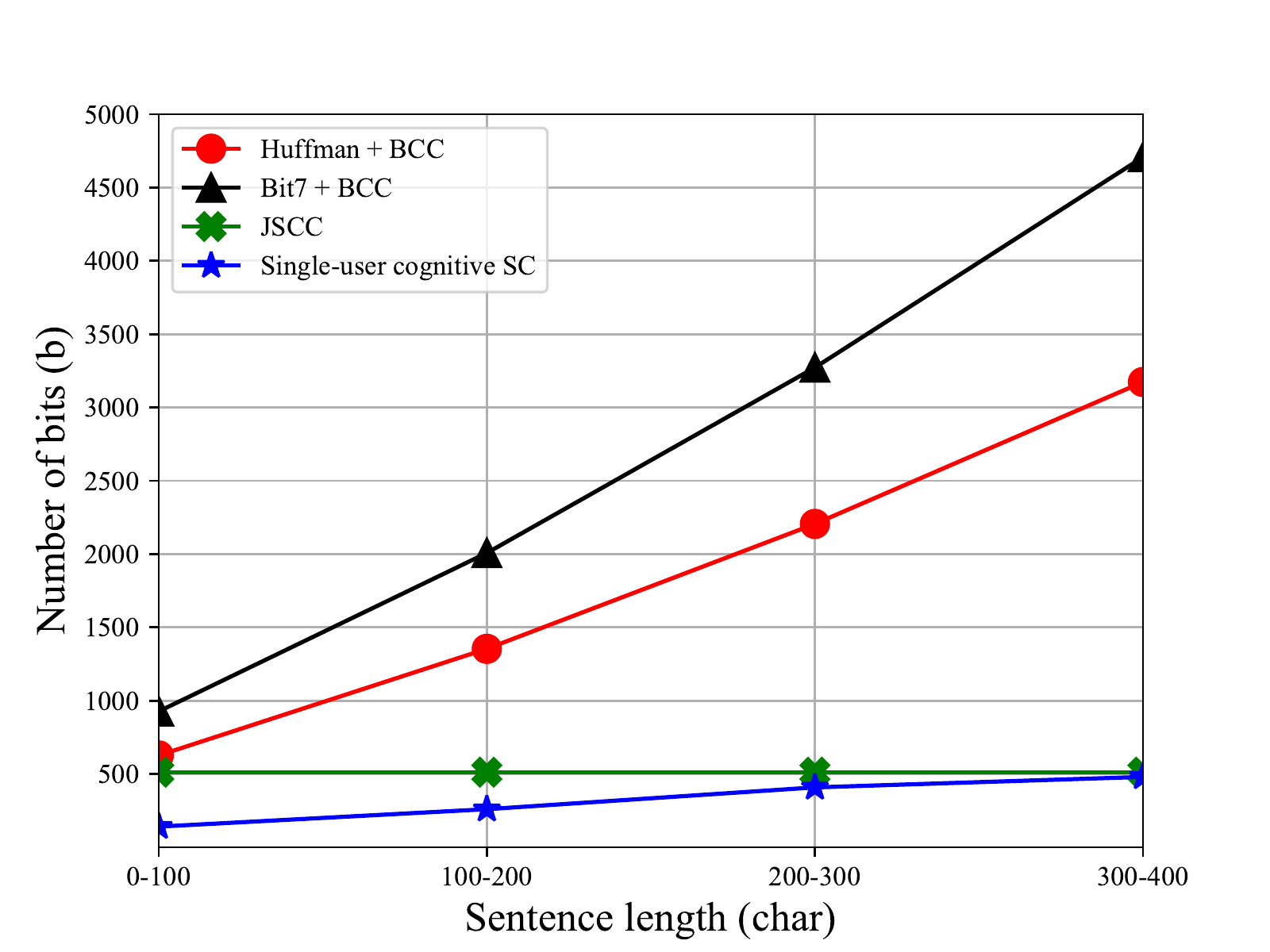}
}
\subfigure[] {
    \label{fig:b}
    \includegraphics[width=2.5 in]{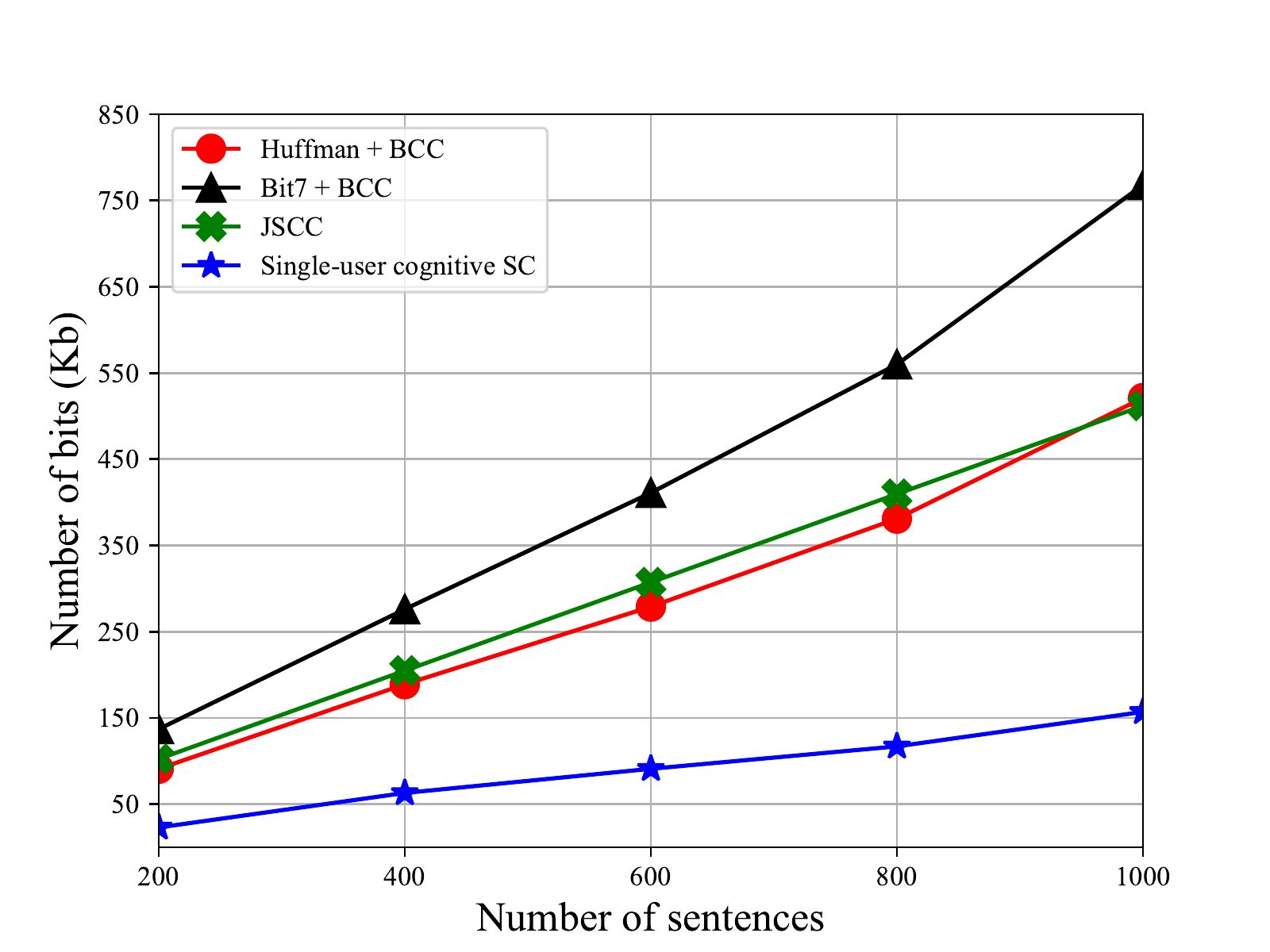}
}
\caption{Comparison of the number of bits required to be transmitted by using different schemes} 
\label{fig:6}  
\end{figure}

Fig. 6(a) shows the number of bits required to be transmitted versus the sentence length. The sentence 
length depends on the number of characters. It is clear that our proposed single-user cognitive semantic 
communication system achieves the largest text compression rate among the benchmark systems, especially 
for long sentences. The reason is that our proposed system can capture the semantic equivalence and 
realize the semantic compression. Moreover, the knowledge graph used as the sharing knowledge base 
between the transmitter and the receiver further improves the coding efficiency. Furthermore, 
since a message can be mapped to one or more triplets by our proposed Text2KG technology, a variable bit 
length coding is realized in our proposed single-user cognitive semantic communication system. 
Compared with the fixed bit length coding approach, such as JSCC, our proposed framework exploits a 
variable bit length coding and improves reasonableness and efficiency. Fig. 6(b) shows the number of bits required to be transmitted versus the number of texts. 
As shown in Fig. 6(b), the proposed single-user cognitive semantic 
communication system transmits the same number of texts with less data compared with all benchmark 
systems. It further indicates that our proposed single-user cognitive semantic communication system 
has a superior compression rate.

\begin{figure}
\centering
\subfigure[] { \label{fig:a}
\includegraphics[width=2.5 in]{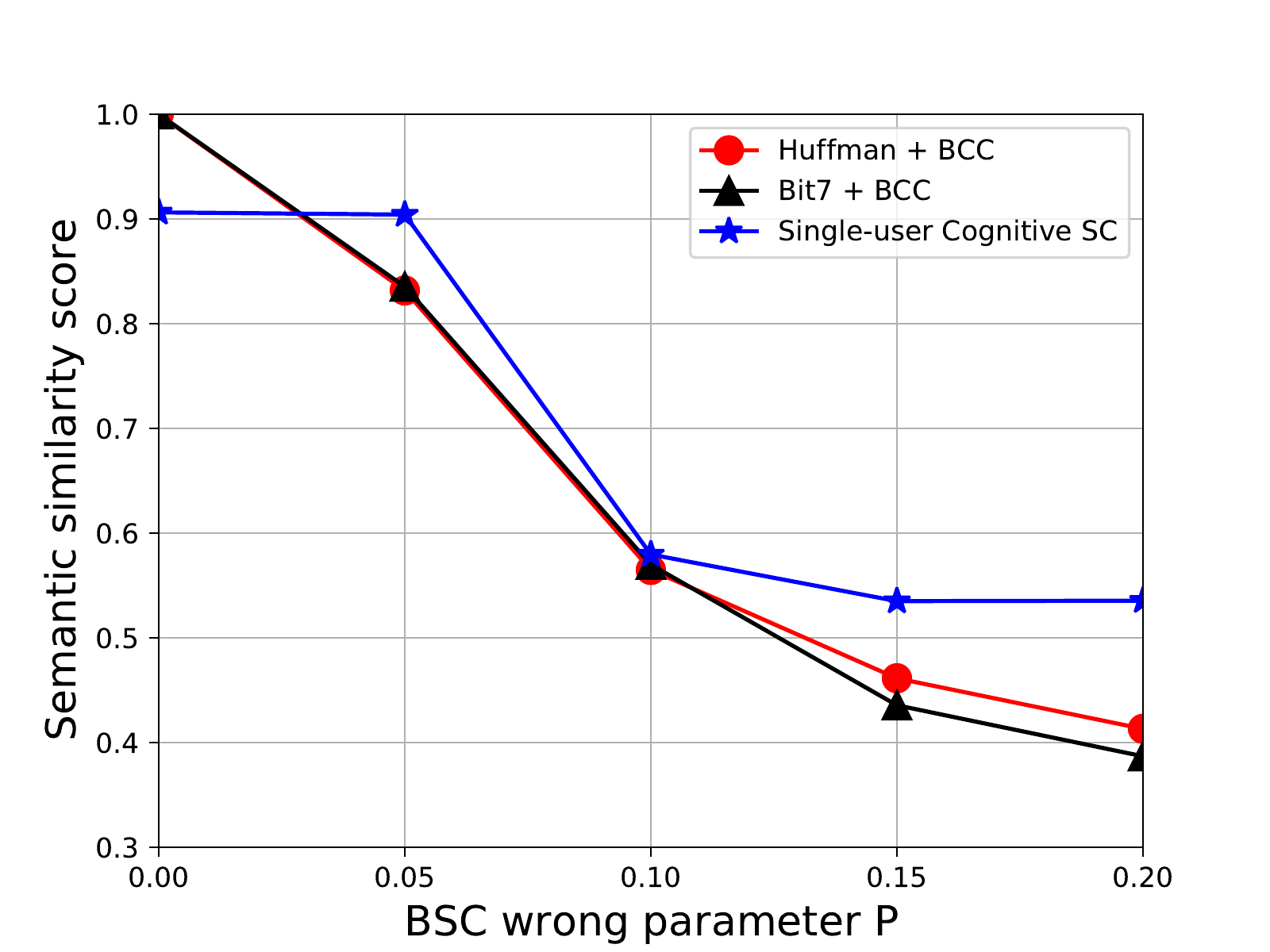}
}
\subfigure[] { \label{fig:b}
\includegraphics[width=2.5 in]{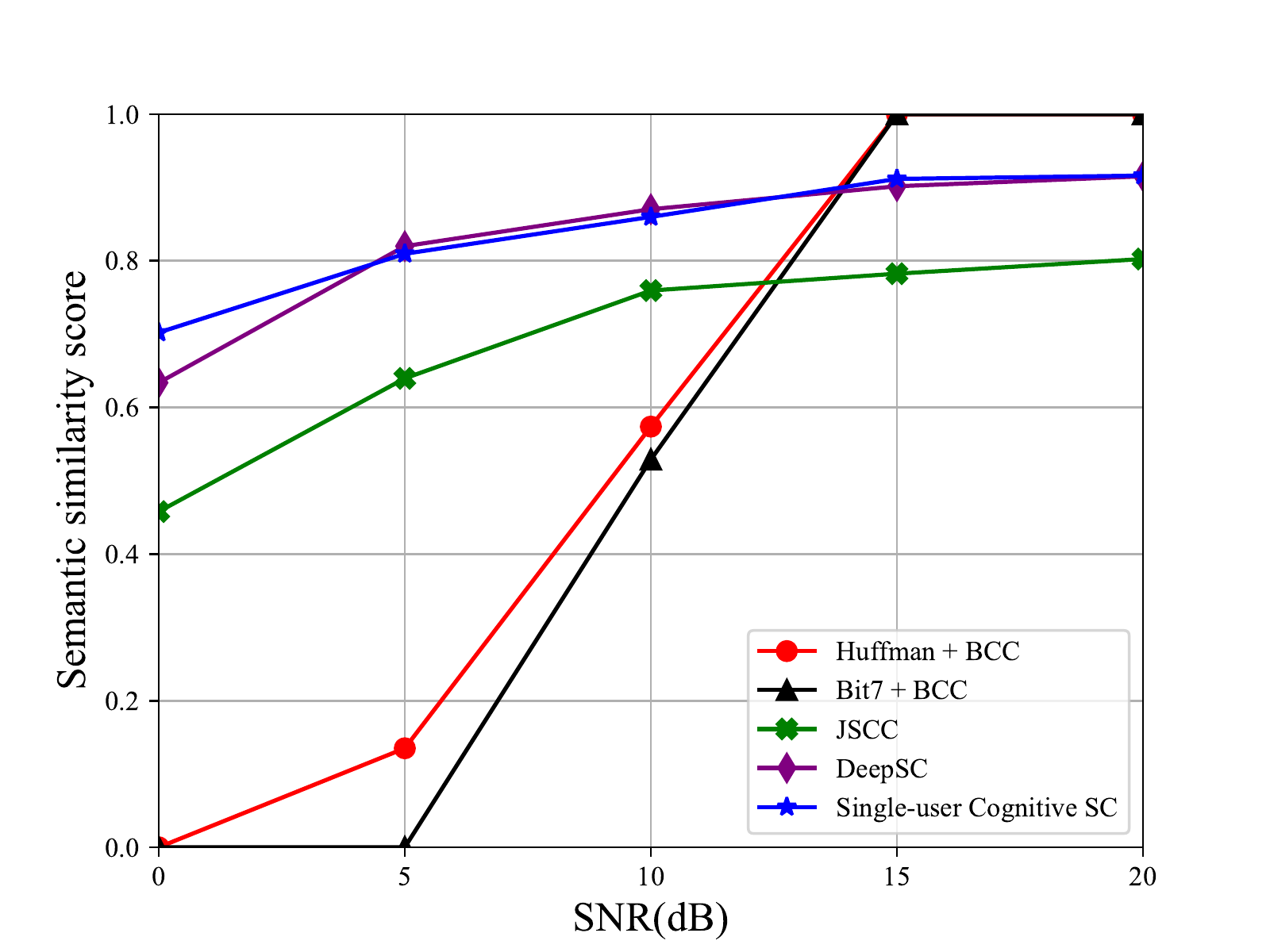}
}
\subfigure[] { \label{fig:c}
\includegraphics[width=2.5 in]{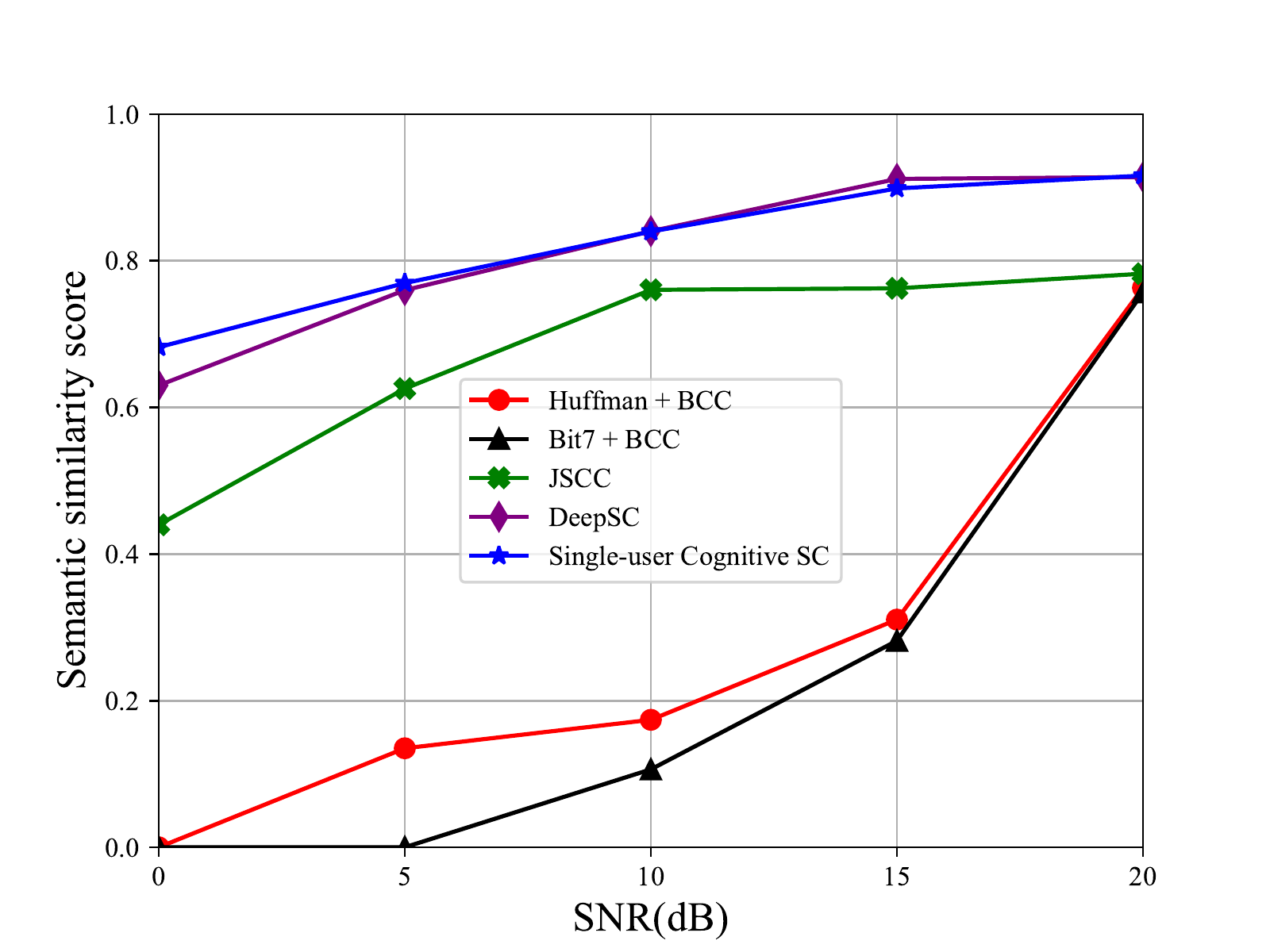}
}
\subfigure[] { \label{fig:d}
\includegraphics[width=3 in]{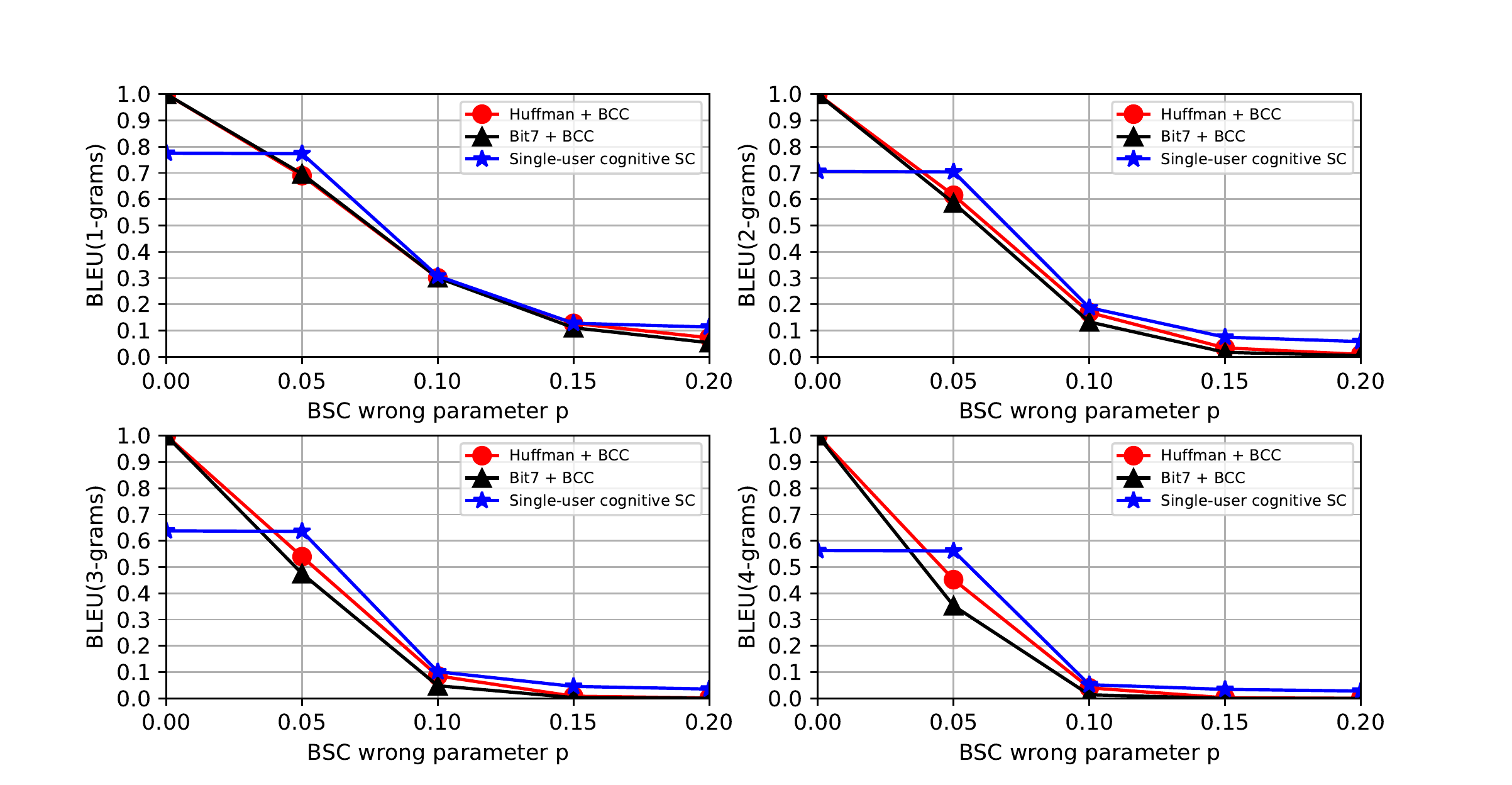}
}
\caption{Comparison of semantic similarity score and BLEU score by using different single-user schemes under difference channels.}
\label{fig.7}
\end{figure}

Fig. 7(a) shows the semantic similarity score versus the BSC wrong parameter $p$ achieved by using our 
proposed single-user cognitive semantic communication system and the benchmark systems. The semantic 
similarity score is used to measure the semantic recovery performance. Note that $p$ represents the 
channel wrong parameter and a larger $p$ represents a worse channel environment. As shown in Fig. 7(a), 
the semantic similarity scores of the benchmark systems are all close to 1 and higher than our proposed 
system when $p<0.05$. Since the channel decoding corrects the errors that occur during transmission, 
the benchmark systems can recover the text without any loss at the receiver. However, the semantic 
information detection in our system has data loss which is actually the reason that our system can 
transmit less data with high semantic fidelity. Meanwhile, the semantic similarity score of our system 
is also higher than 0.9, which means that the messages can be well understood. The relationship of the 
semantic similarity score with how well a message can be understood is showed in Table VIII. When $p>0.05$, the channel 
environment is poor, the channel coding cannot correct all errors. In this case, the performance of the 
benchmark systems degrades rapidly. In contrast, since our proposed cognitive semantic communication 
system can more effectively capture and recover semantic information and correct errors, it can generate 
easy-to-understand messages. 

\begin{table}
    \normalsize
    \centering
    \renewcommand\arraystretch{0.9}
    \setlength{\tabcolsep}{10mm}
    \newcommand{\tabincell}[2]{\begin{tabular}{{@{}#1@{}}#2\end{tabular}}}
    \resizebox{\textwidth}{!}{
        \begin{tabular}{|c|c|}
        \hline  
        Semantic Similarity Scores&Examples\\ \hline 					
        1.0 & TX: Peter Creamer was born in Hartlepool. \\ & RX: Peter Creamer was born in Hartlepool.\\ \hline
        0.9 & TX: David Tong is affiliated with Blackpool F.C.\\ & RX: David Tong plays for Blackpool F.C.\\ \hline
        0.7 & TX: George Mallia plays for the Malta national football team.\\
        & RX: Mallia plays for the England national under-21 football team.\\ \hline
        0.5 & TX: Chris Allen (footballer) was born in 1972.\\ & RX: Chris Allen is affiliated with Queens Park Rangers F.C.\\ \hline
        0.1 & TX:Josh Smith plays for Trinity University in Texas. \\ & RX: rt Jay iith cePnmfdo lrotltwar vivergelLfnt imaeli.\\ \hline
    \end{tabular}
    }
    \caption{The relationship of the semantic similarity scores with how well a message can be understood.}
\end{table}

Fig. 7(b) shows the semantic similarity score versus the signal-to-noise ratio (SNR) achieved by using the 
proposed single-user cognitive semantic communication system and the benchmark systems under the AWGN channels. 
Note that for the traditional methods, i.e., Huffman with BCC coding and Bit7 with BCC coding,  
there are almost no errors during transmission when SNR \textgreater 15 dB. Thus, the sentence similarity score of the traditional 
methods in Fig. 7(b) almost converges to 1. However, with the reduction of SNR, the performance of the
traditional methods is significantly degraded and is obviously poorer than that of the semantic communication systems when SNR \textless 10 dB. 
Our proposed cognitive semantic communication system shows the same behavior as the DNN-based JSCC method and DeepSC in Fig. 7(b). 
In particular, the performance of our proposed system is close to that of the DeepSC. 
However, since the semantic errors can be corrected by inference modules, it is clear that our proposed single-user cognitive 
communication system is more competitive than the DeepSC in the low SNR regime. 
Fig. 7(c) shows the semantic similarity score versus the SNR achieved by using our 
proposed single-user cognitive semantic communication system and the benchmark systems under Rayleigh fading channels.
Fig. 7(c) exhibits the same behavior as that of Fig. 7(b). The semantic communication systems, including DeepSC, JSCC and our proposed
single-user cognitive semantic communication system, also achieve much higher score than the traditional systems
in terms of the semantic similarity. clearly, our proposed system achieves better performance when SNR \textless 5 dB under Rayleigh fading channels.
Fig. 7(d) shows the same tendency with 
the BLEU score as the performance criterion. It is clear that our proposed system is more competitive and robust in poor channel 
environments. A few cases between the transmitter and the receiver are shown in Table IX in order 
to demonstrate the efficiency of our proposed single-user cognitive semantic communication system.
\begin{figure}[!t]
    \centering
    \subfigure[] { \label{fig:}
    \includegraphics[width=2.5 in]{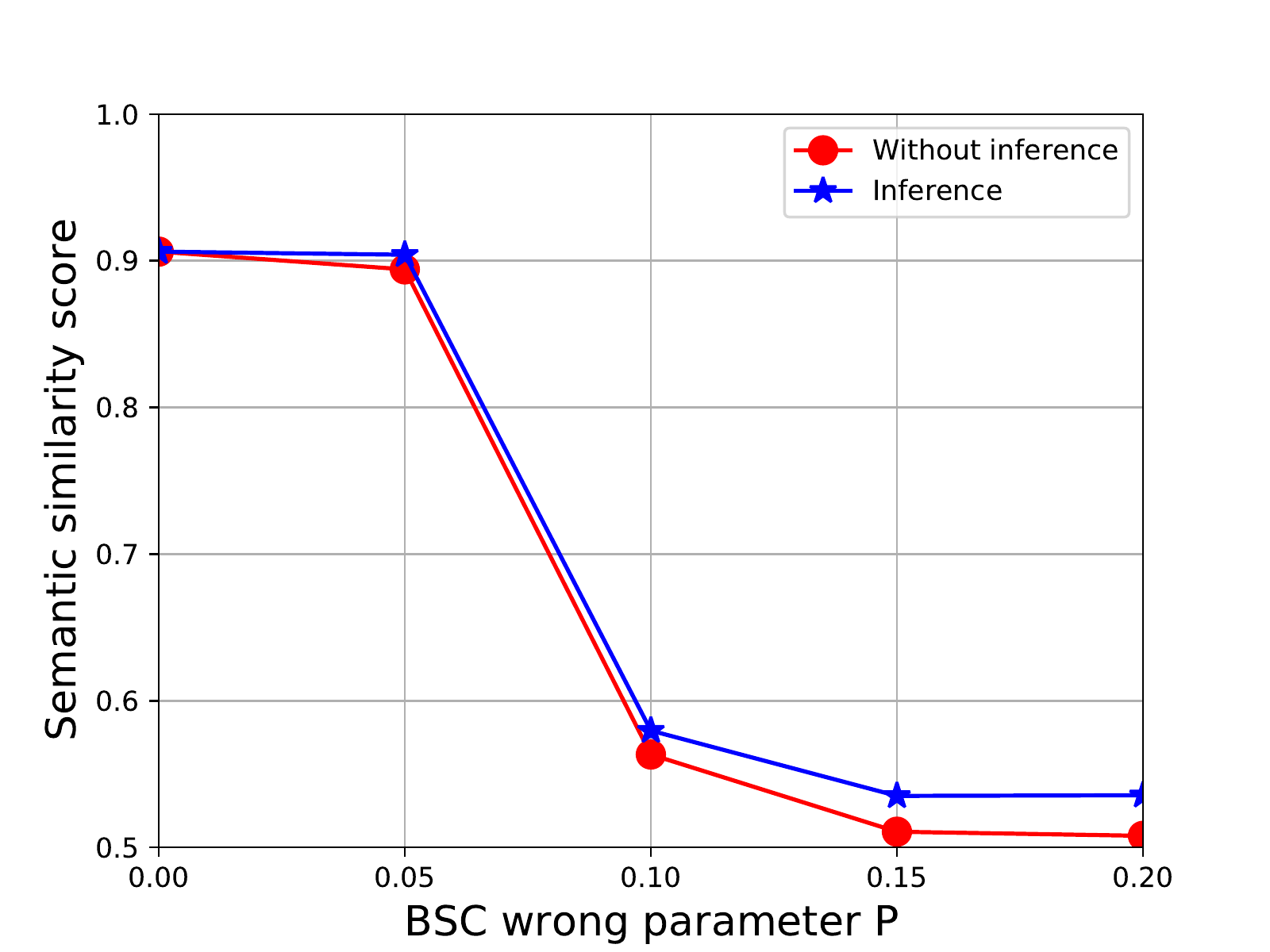}
    }
    \subfigure[]{  \label{fig:b}
    \includegraphics[width=2.5 in]{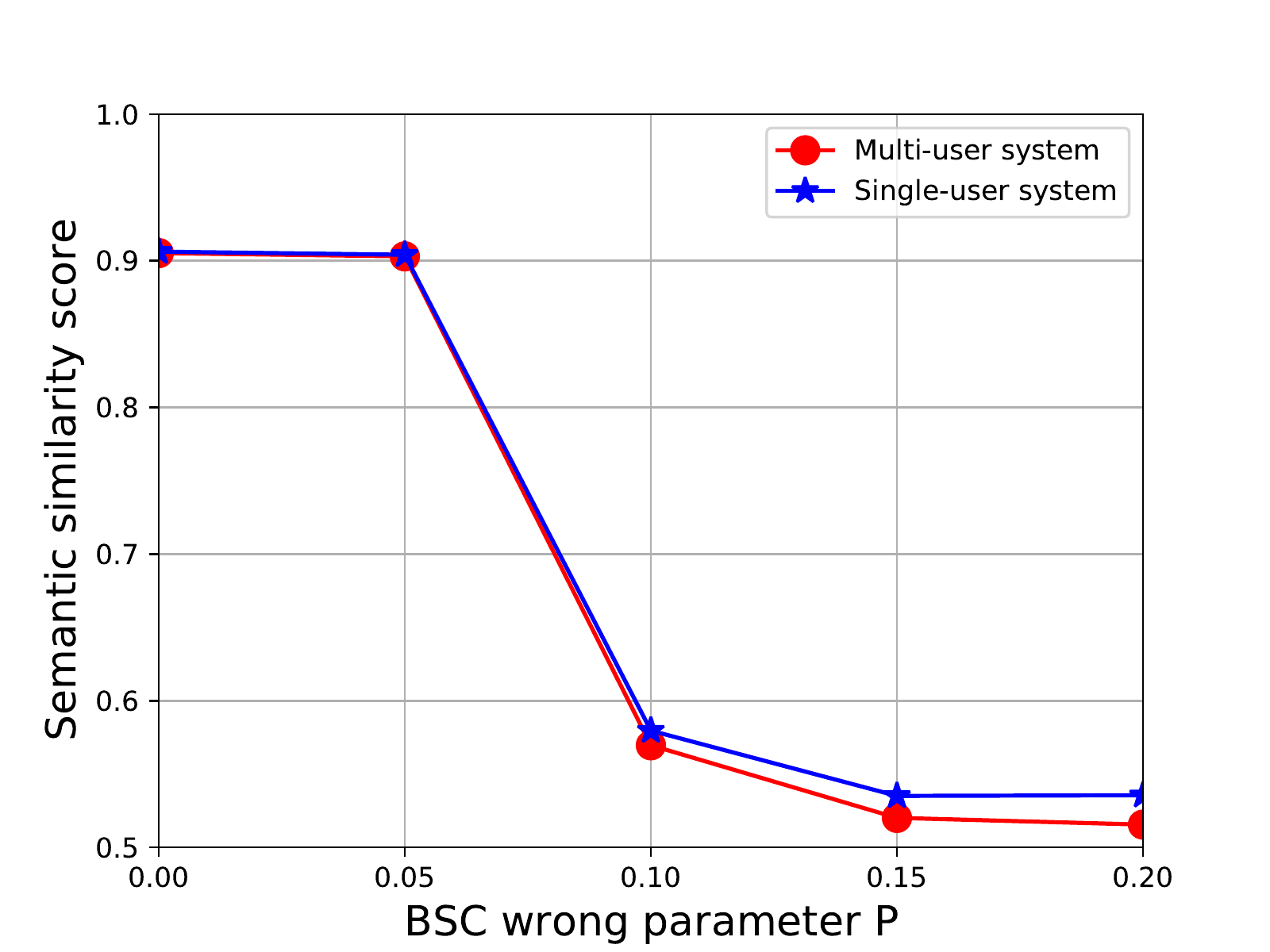}
    }
    \caption{(a) Comparison of semantic similarity scores by using different multi-user schemes under difference channels.
    (b) The comparison of the semantic similarity scores between our proposed multi-user and the single-user cognitive semantic communication systems.} \label{fig.6}
\end{figure}

Fig. 8(a) shows the effect of the inference model on the semantic similarity score. As shown in Fig. 8(a), 
the semantic similarity scores of the system that has the inference module is very close to the system 
that does not have it when $p<0.1$. The reason is that the channel decoding corrects most errors that 
occur during transmission, and the triplets reconstructed at the destination are always true and the 
inference module does not work. On the contrast, the semantic similarity score of the system with 
inference is always higher than that of the system without it when $p >=0.1$. The reason is that since 
the channel decoding fails to work, the partial triplets reconstructed at the destination are wrong. 
In this case, our proposed correction algorithm can correct the incorrect triplets based on the inference 
rules of the knowledge graph embedding model.

\begin{table*}
    \normalsize
    \centering
    \renewcommand\arraystretch{0.9}
    \setlength{\tabcolsep}{10mm}
    \newcommand{\tabincell}[2]{\begin{tabular}{{@{}#1@{}}#2\end{tabular}}}
    \resizebox{\textwidth}{!}{
    \begin{tabular}{|c|c|}
    \hline 					
    Lossless & TX: Boo Young-tae plays for Yangju Citizen FC. \\ & RX: Boo Young-tae plays for Yangju Citizen FC.\\
    \hline
    Rephrasing but lossless semantics &   TX: David Tong \emph{\textbf{is affiliated with}} Blackpool F.C.\\ & RX: David Tong \emph{\textbf{plays for}} Blackpool F.C.\\
    \hline
    Align synonyms & TX: The University of Tampa is located in \emph{\textbf{the United States}}.\\
    & RX: The University of Tampa is located in \emph{\textbf{the US}}.\\
    \hline
    An inexplicable error & TX: Sidney Govou plays for \emph{\textbf{Olympique Lyonnais}}.\\ & RX: Sidney Govou plays for \emph{\textbf{RC Strasbourg}}.\\
    \hline
    Disambiguation & TX: Batchoy includes \emph{\textbf{chicken}}.\\ &RX: \emph{\textbf{Chicken}} is an ingredient of Batchoy.\\
    \hline
    \end{tabular}
    }
    \caption{Cases which are transmitted and received using our proposed system when $p=0.1$.}
\end{table*}

\begin{figure}[!t]
\centering
\subfigure[] { \label{fig:}
\includegraphics[width=2.5 in]{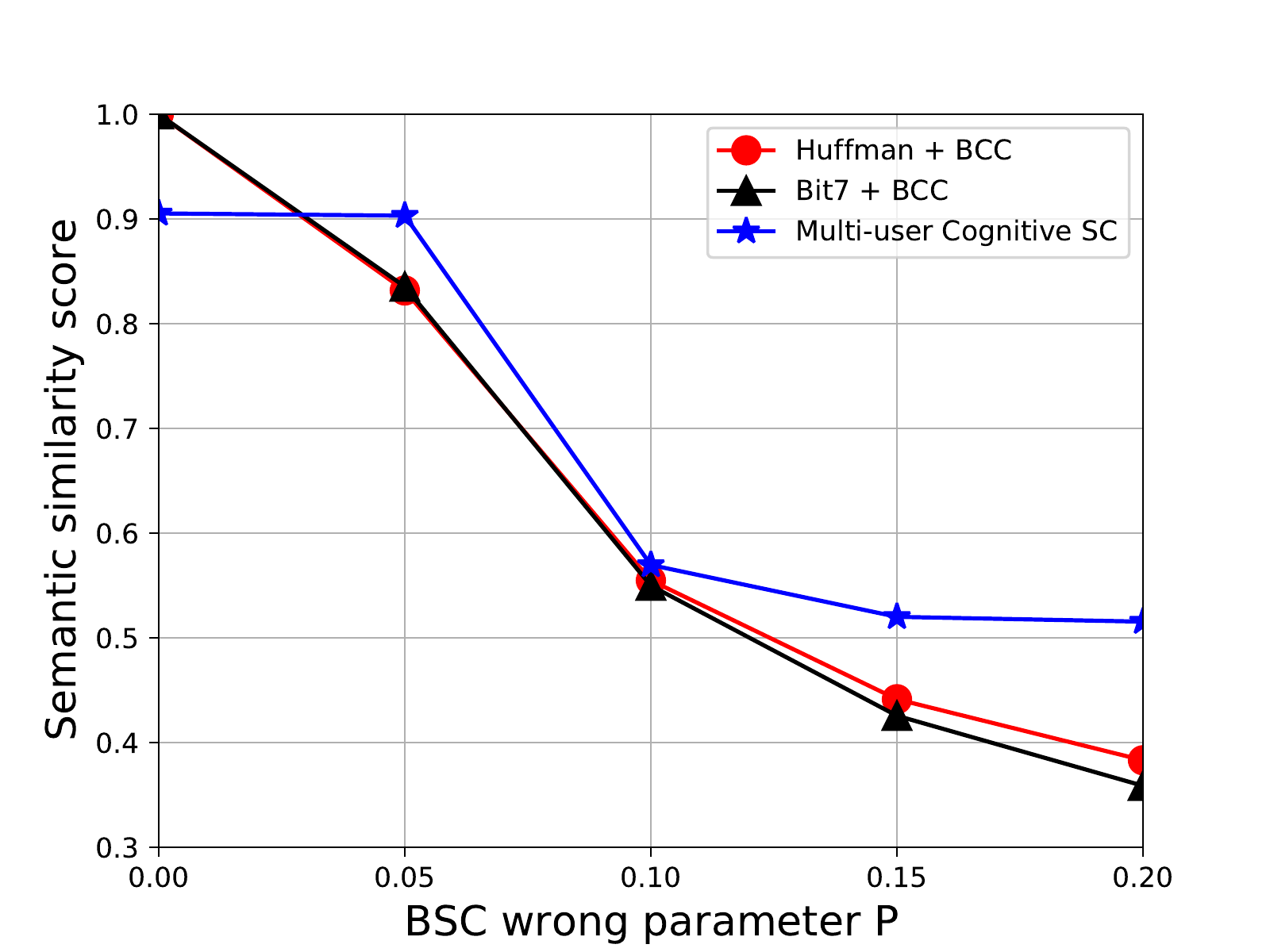}
}
\subfigure[]{  \label{fig:b}
\includegraphics[width=2.5 in]{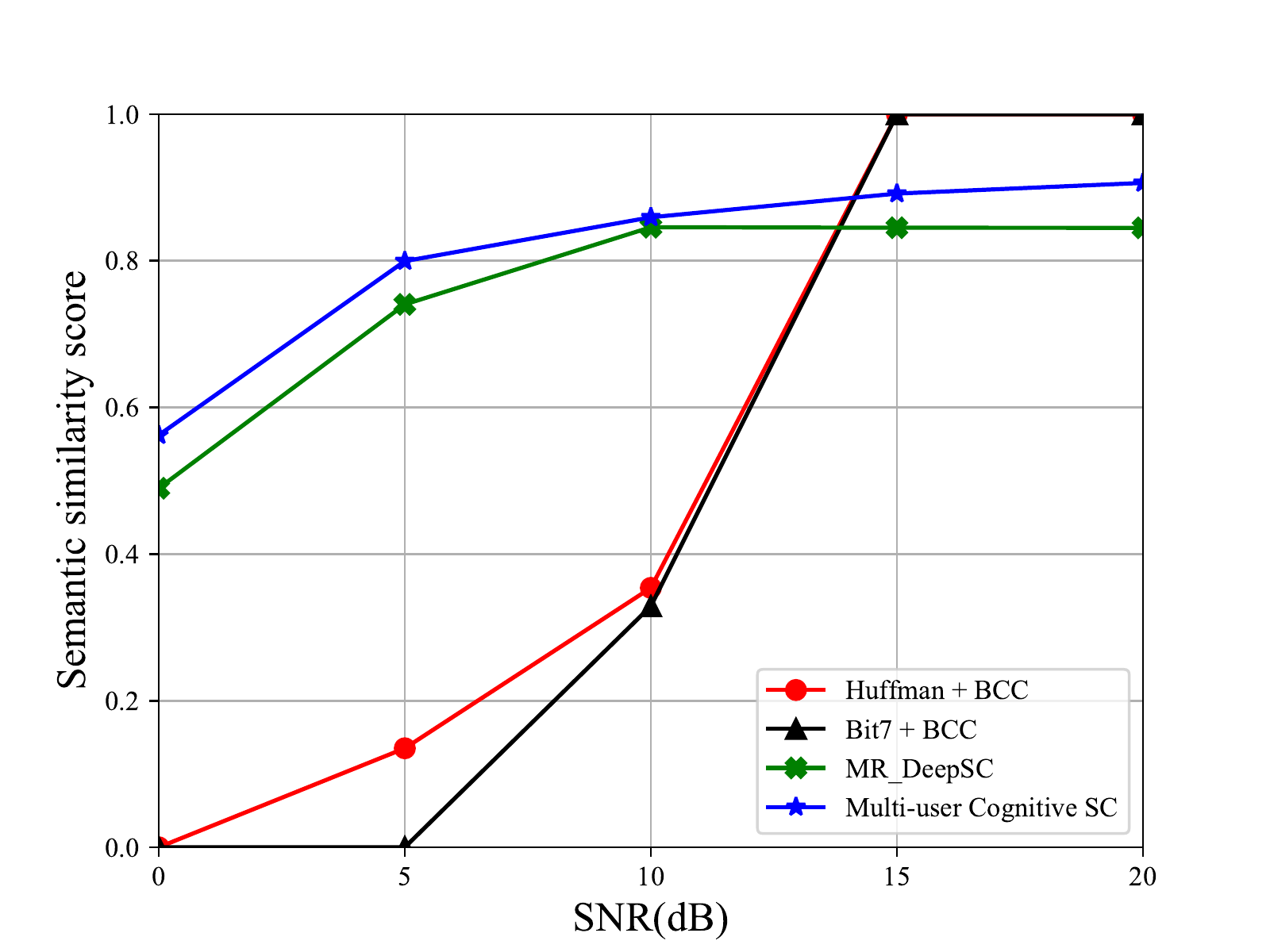}
}
\subfigure[]{  \label{fig:c}
\includegraphics[width=2.5 in]{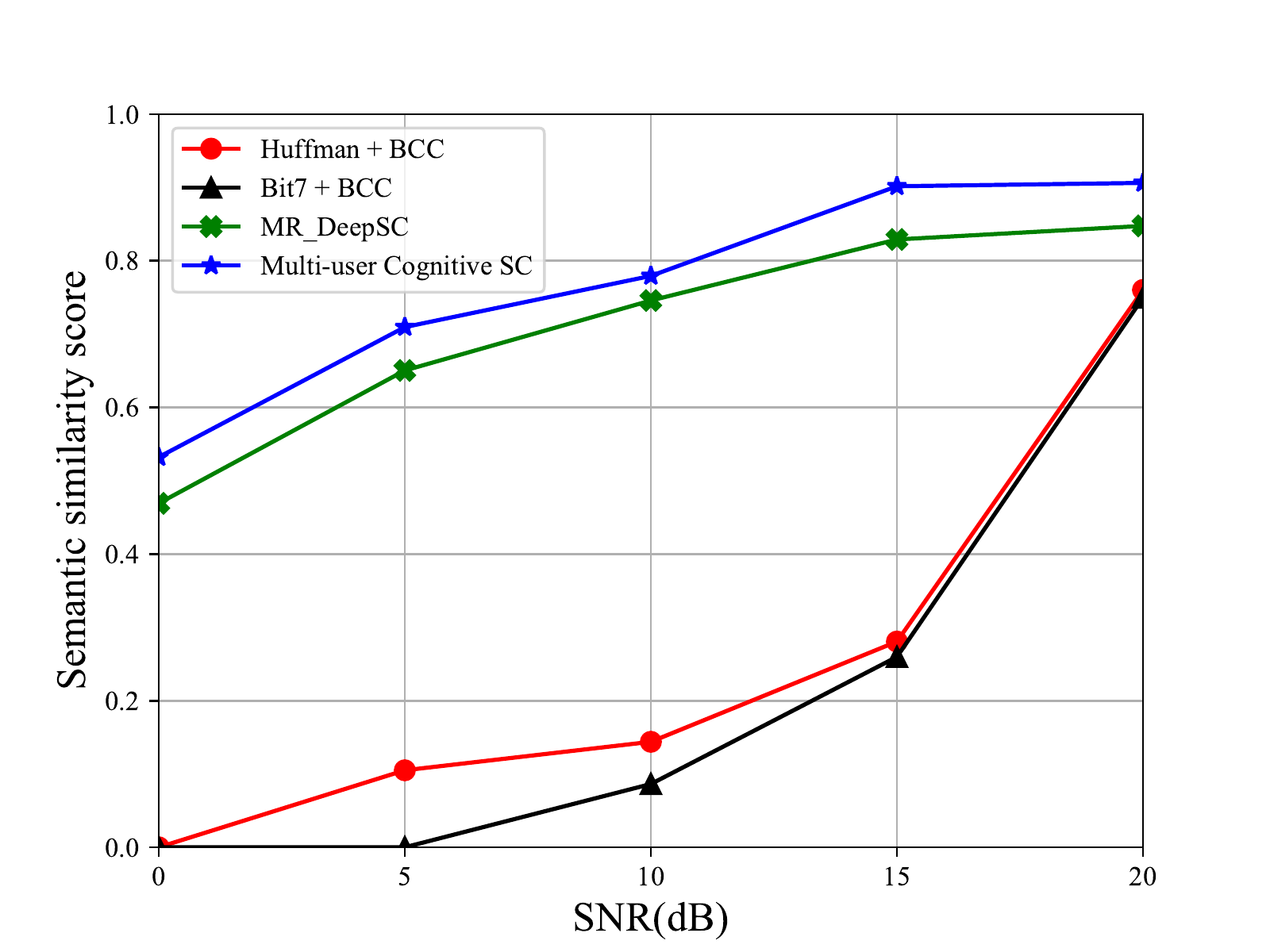}
}
\caption{Comparison of semantic similarity scores by using different multi-user schemes under difference channels.} \label{fig.6}
\end{figure}
Fig. 9(a) shows the semantic similarity score versus the BSC wrong parameter $p$ achieved by using our 
proposed multi-user cognitive semantic communication system. The dataset and simulation setup of the 
multi-user system is the same as those used for the single-user system. Note that the simulation of our 
multi-user cognitive semantic communication system consists of two transmitters and two receivers. It can 
be seen that, similar to the single-user cognitive semantic communication system, the semantic similarity 
score of the benchmark systems is higher than that of our proposed multi-user system and is close to 1 
when $p<0.05$. The reason is that the messages transmitted at the source and reconstructed at the 
destination are identical through the benchmark systems when channel environment is good. However, the 
messages transmitted at the source and reconstructed at the destination are semantic equivalent but have 
different structures through our proposed multi-user cognitive semantic communication system. Meanwhile, 
the semantic similarity score of our proposed cognitive multi-user semantic communication system is also 
higher than 0.9, which means that the semantic information contained messages is transmitted successfully. 
And this behavior is in accordance with the simulation results in \cite{xie2021task-O}. When $p>0.05$, 
it shows the same tendency as that in the single-user system. It is clear that our proposed multi-user 
cognitive semantic communication system is more competitive and robust in poor channel environments. 

Fig. 9(b) shows the semantic similarity score versus the SNR achieved by using our 
proposed multi-user cognitive semantic communication system and the benchmark systems under AWGN channels. 
In the comparison of the semantic communication systems with the traditional communication systems, it is observed that
the performance and stability of all the semantic communication systems at low SNR are significantly better than those of the traditional communication systems.
Moreover, it is clear that our proposed multi-user cognitive semantic communication achieve better performance 
than MR-DeepSC \cite{hu2022one}. The reasons are as follows. On the one hand, the semantic correction
algorithm is exploited to correct the semantic errors occurring at the transmission and improves the robustness of our proposed communication system.
On the other hand, a semantic recognizer based on the DistilBERT is built to distinguish different users according to the different emotions \cite{hu2022one}.
Different users have their own private knowledge graphs.
A message recovery algorithm is exploited to distinguish messages of different users by matching the knowledge
level to avoid the situation where the emotions of some messages are ambiguous. 
Fig. 9(c) shows the semantic similarity score versus the SNR achieved by using the 
proposed multi-user cognitive semantic communication system and the benchmark systems under Rayleigh fading channels.
Fig. 9(c) and Fig. 9(b) show the same behavior. Clearly, our proposed multi-user cognitive semantic communication systems are robust to different channel conditions. 

Fig. 8(b) shows the comparison of the semantic similarity score achieved by using the multi-user cognitive 
semantic communication system with that obtained by using the single-user cognitive semantic communication 
system. As shown in Fig. 8(b), the semantic similarity score of the multi-user cognitive semantic communication 
system and that of the single-user cognitive semantic communication system are close when $p<0.1$. However, 
the semantic similarity score of the multi-user cognitive semantic communication system is lower than that 
of the single-user cognitive semantic communication system when $p>0.1$. The reason is that our proposed 
multi-user cognitive semantic communication system distinguishes different users based on their own knowledge 
graphs. When the channel environment is poor, the reconstructed messages at the destination can be wrong 
and cannot be matched with the true user's knowledge graph. Thus, they may not be assigned to the true user. 
In addition, most of the reconstructed messages are correct when the channel environment is good. In this 
case, the results occurring under the poor channel environment do not happen. This behavior is consistent 
with the simulation results shown in \cite{xie2021task-O}.

The computational complexities of the proposed single-user cognitive semantic communication system, the DeepSC \cite{xie2021deep}, the JSCC \cite{8461983}, the Huffman coding, are compared
in Table X in terms of the average processing runtime per sentence. All simulations are performed with AMD EPYC 7302@3.0GHz and NVIDIA GeForce GTX 3090.
The traditional system, such as Huffman, has lower runtime than the DNN-based systems. The JSCC requires
a relatively low runtime due to its simple network architecture. Correspondingly, it has poorer semantic processing capability. 
Moreover, the runtime of the DeepSC is slightly higher than that of JSCC but with significant performance improvement.  
As a comparison, our proposed single-user cognitive semantic communication system requires slightly higher runtime than DeepSC. The reason is 
that the computational complexities of our proposed semantic alignment algorithm and semantic correction algorithm
increase with the size of knowledge graph.
\begin{table}
    \normalsize\centering
    \begin{center}
    \renewcommand\arraystretch{0.78}
    \setlength{\tabcolsep}{0.7mm}
    \newcommand{\tabincell}[2]{\begin{tabular}{@{}#1@{}}#2\end{tabular}}
    \begin{tabular}{c|c|c|c|c}
    \Xhline{2pt}
    Models & Our proposed system & DeepSC \cite{xie2021deep} & JSCC \cite{8461983} & Huffman \\
    \Xhline{1pt}
    Runtime & 4.84ms & 3.07ms & 2.28ms & 2.14ms \\
    \Xhline{2pt}
    \end{tabular}

\caption{The time complexity of all strategies.}
\end{center}
\end{table}
\subsection{The SDR Prototype Systems and Experimental Results}
\begin{figure}
\centering
\subfigure[] { \label{fig:a}
\includegraphics[width=0.4\columnwidth]{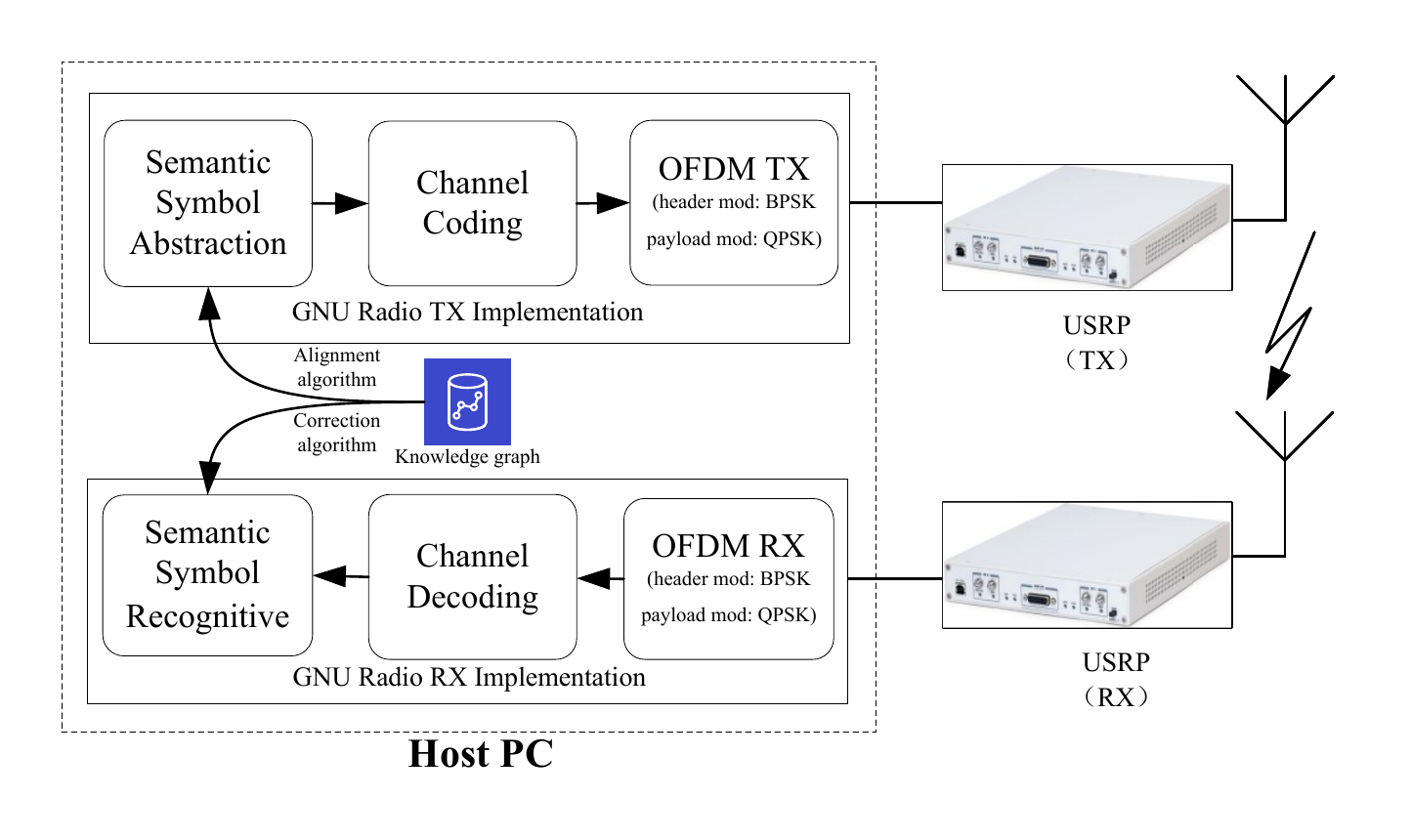}
}
\subfigure[] { \label{fig:b}
\includegraphics[width=0.3\columnwidth]{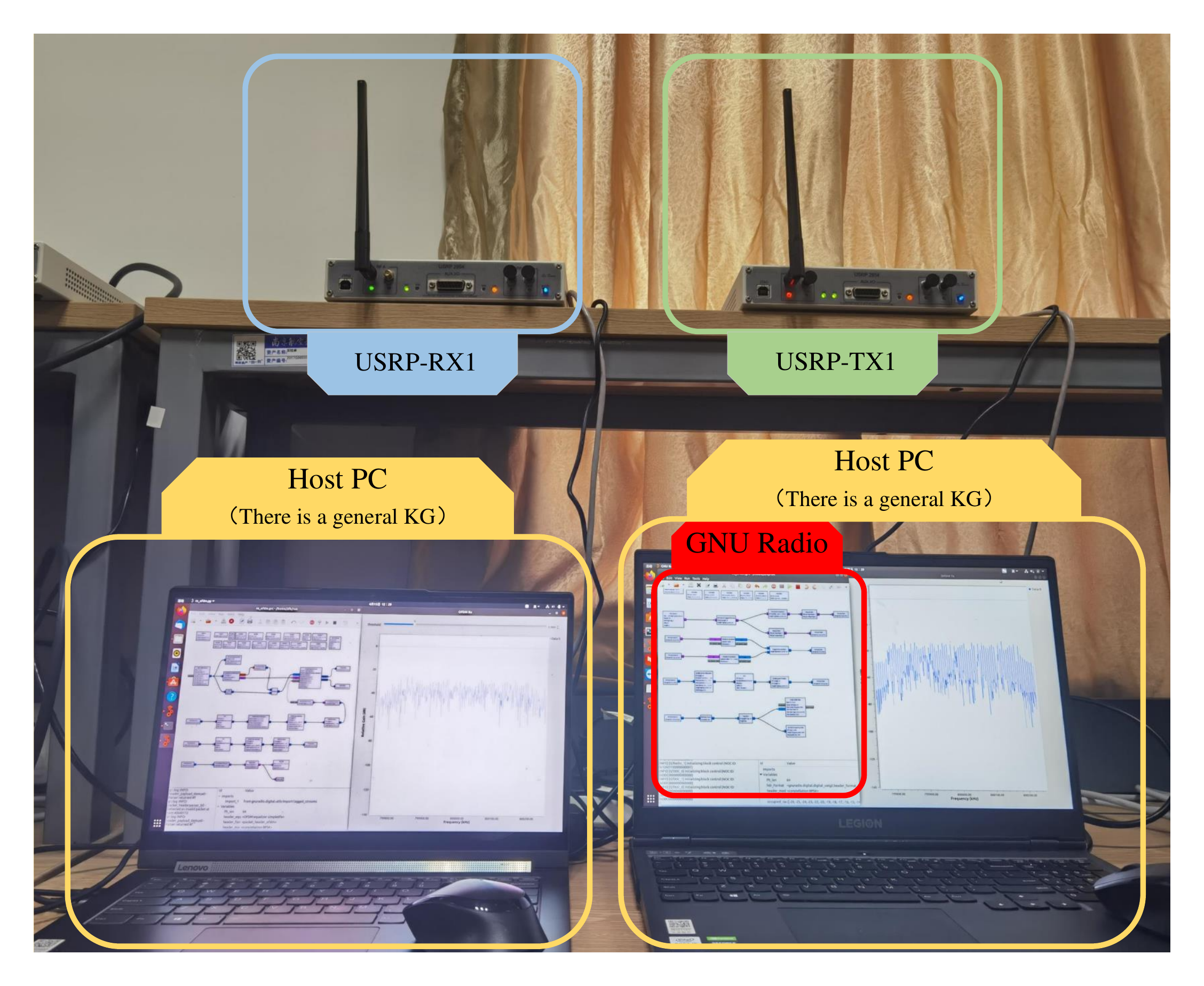}
}
\subfigure[] { \label{fig:c}
\includegraphics[width=0.7\columnwidth]{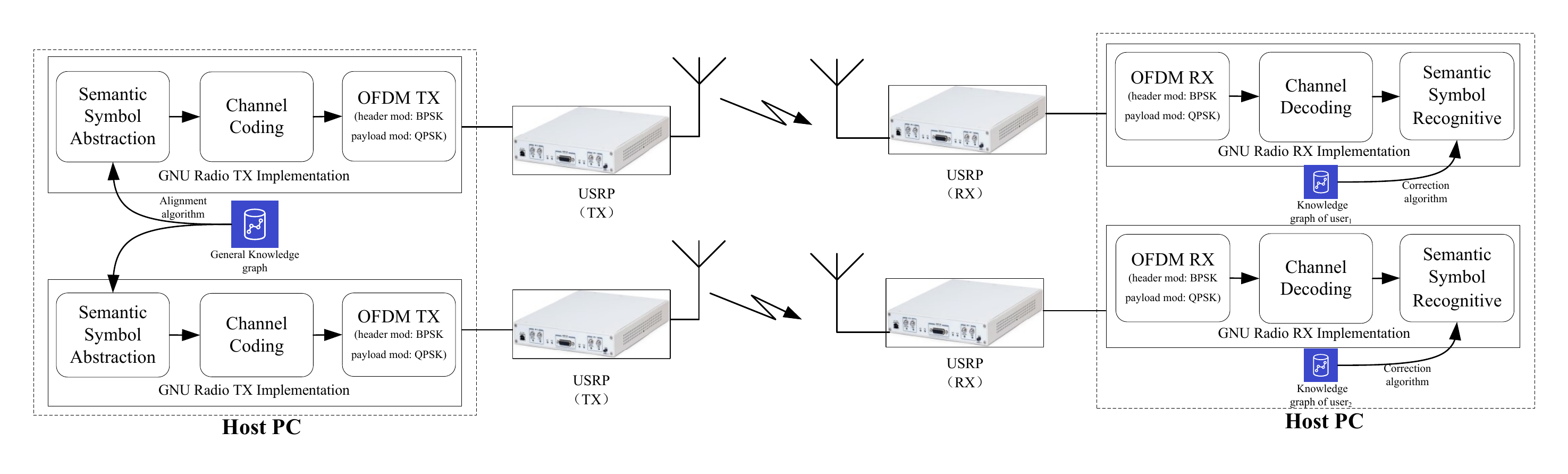}
}
\subfigure[] { \label{fig:d}
\includegraphics[width=0.7\columnwidth]{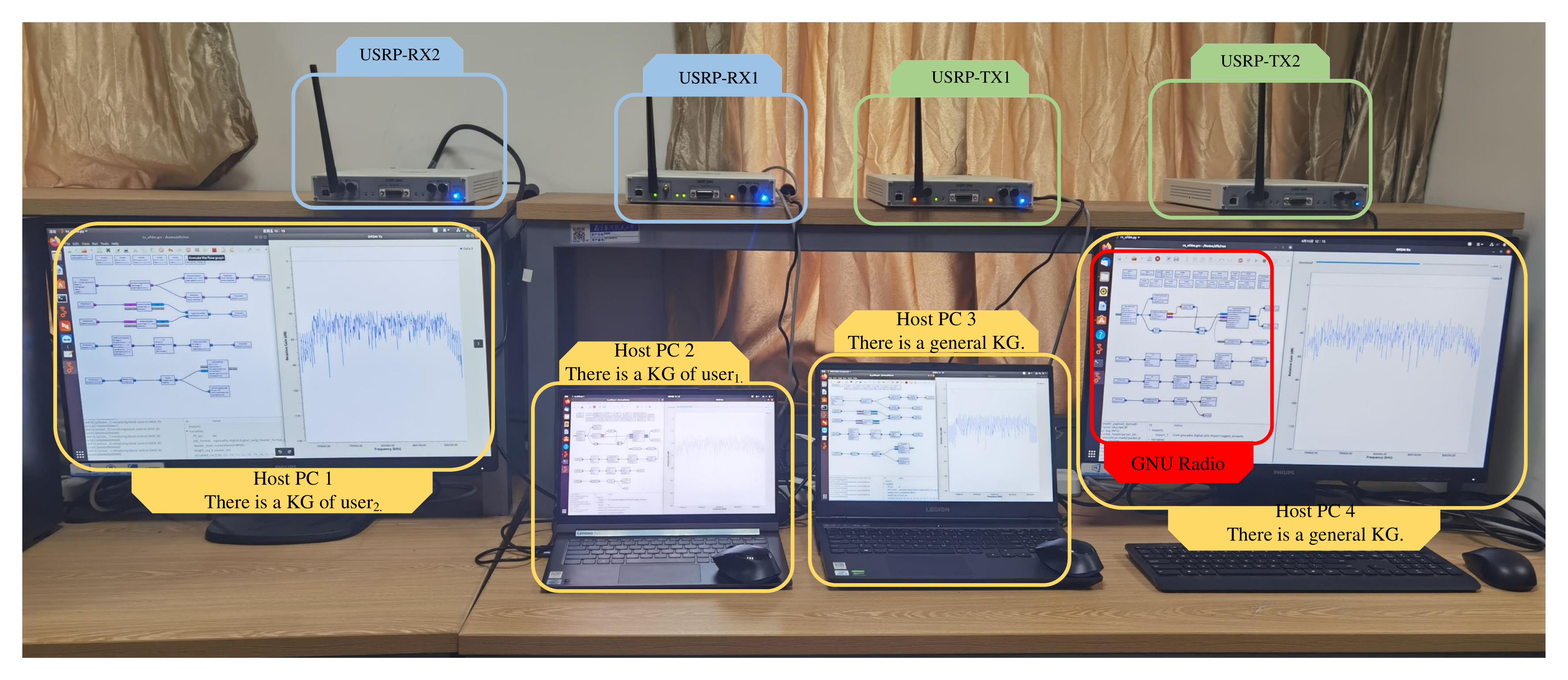}
}
\caption{The hardware-in-the-loop simulation block diagram and the photograph of our proposed single-user cognitive semantic communication system is shown in (a), (b), respectively. The hardware-in-the-loop simulation block diagram and the photograph of our proposed multi-user cognitive semantic communication systems are shown in (c), (d), respectively.}
\label{fig}
\end{figure}
In order to demonstrate the feasibility of our proposed cognitive semantic 
communication systems, an experimental SDR testbed is built to implement our 
proposed cognitive semantic communication systems. The hardware-in-the-loop 
simulation block diagram and photographs of the overall experimental system 
are shown in Fig. 10.
\subsubsection{The SDR Prototype Systems}
To implement the SDR testbed, we deploy four universal software radio peripheral (USRP) 2954R with the full signal processing handled by GNU Radio which is a free open-source software platform installed in personal computer and USRP hardware drivers (UHD). The target monitoring frequency band of USRP-2954R is 10MHz-6GHz with a maximum instantaneous real-time bandwidth of 160MHz and a maximum I/Q sampling rate relatively high as 200 MS/s. In our experiment, two desktop computers are connected to two USRPs via MXI Express four-lane cables. Moreover, two gigabit Ethernet (GigE) connections are established between two other USRPs and two laptop computers, using the USRP SFP+ port. The operating systems are Ubuntu 20.04 LTS with kernel version 5.8.0-60-generic, installed with UHD 3.15.0 and GNU Radio 3.8.3.
\subsubsection{Experimental Results}
In our experiment, 50 sentences are transmitted by using our proposed 
single-user cognitive semantic communication system and traditional 
communication systems on our experimental SDR testbed. The traditional communication systems are realized by
the separate source and channel coding technologies. Specifically, the channel 
coding scheme of our proposed system is the same as that of the benchmark 
communication systems, namely, the binary convolutional codes. Two source coding schemes are exploited for the benchmark
systems, namely, Huffman and fixed-length coding. In Fig. 11(a) and Fig. 11(b), in the comparison of the proposed cognitive semantic 
communication system with the traditional communication systems, it is 
observed that the proposed cognitive semantic communication system has 
higher compression rate. In parallel, it can be seen that the semantic 
similarity score of the our proposed system and traditional communication 
systems are all higher than 0.8. Hence, the decoded sentences at the 
receiver by our proposed cognitive semantic communication system and 
the traditional systems can be well understood.

In order to demonstrate the process of our proposed semantic communication 
system clearly, a sentence is token as an example. The sentence \lq Chatou is located in France\rq \ is transmitted by our proposed 
single-user cognitive semantic communication system. It is aligned to a triplet (\lq Chatou\rq, \lq 
isLocatedIn\rq, \lq France\rq) by our proposed semantic aligned algorithm. After channel coding and 
modulation, it is transmitted by the USRP. At the receiver, some alternative triplets are obtained 
according to the channel decoding, such as (\lq Quirindi\rq, \lq isAffiliatedTo\rq, \lq 
Fc Barcelona \rq), (\lq Chatou\rq, \lq edited\rq, \lq Mersin\rq) and so on. The true triplet 
(\lq Chatou\rq, \lq isLocatedIn\rq, \lq France\rq) is picked out from them by exploiting our proposed 
semantic correction algorithm. Finally, the reconstructed sentence \lq Chatou is located in France\rq \ is 
obtained at the destination. However, the sentence \lq Chatou is loc1ted in Frkn5e\rq \ is obtained at 
the destination by using the traditional communication system. It is clear that our proposed single-user 
cognitive semantic communication system achieves better performance.

For the experiment of our proposed multi-user cognitive semantic communication system, the first source 
transmits the sentence \lq Chatou is located in France\rq \ and the second source transmits the sentence 
\lq Boo Young tae plays for Yangju Citizen FC\rq. Similar to the single-user cognitive semantic 
communication system, those two sentences are aligned to (\lq Chatou\rq, \lq isLocatedIn\rq, \lq France\rq) 
and (\lq Boo Young tae\rq, \lq playsFor\rq, \lq Yangju Citizen FC\rq), respectively. After channel coding 
and modulation, they are transmitted by the USRP. At the receiver, (\lq Chatou\rq, \lq isLocatedIn\rq, 
\lq France\rq) and (\lq Boo Young tae\rq, \lq playsFor\rq, \lq Yangju Citizen FC\rq) are both obtained at 
two destinations by exploiting our proposed semantic correction algorithm. Then, they are distinguished 
by exploiting our proposed message recovery algorithm. Finally, \lq Chatou is located in France\rq \ is 
reconstructed at the first destination and \lq Boo Young tae plays for Yangju Citizen FC\rq \ is 
reconstructed at the second destination. However, \lq Chatou as located in Fmkns1\rq \ and \lq Boa Yr7ong 
ta pliys for Yasgj Citizen WC\rq \ are obtained, respectively, at two the destinations by using the 
traditional communication system. It is clear that our proposed multi-user cognitive semantic 
communication system also achieves better performance.

\begin{figure}
    \centering
    \subfigure[] {
        \label{fig:a}
        \includegraphics[width=2.5 in]{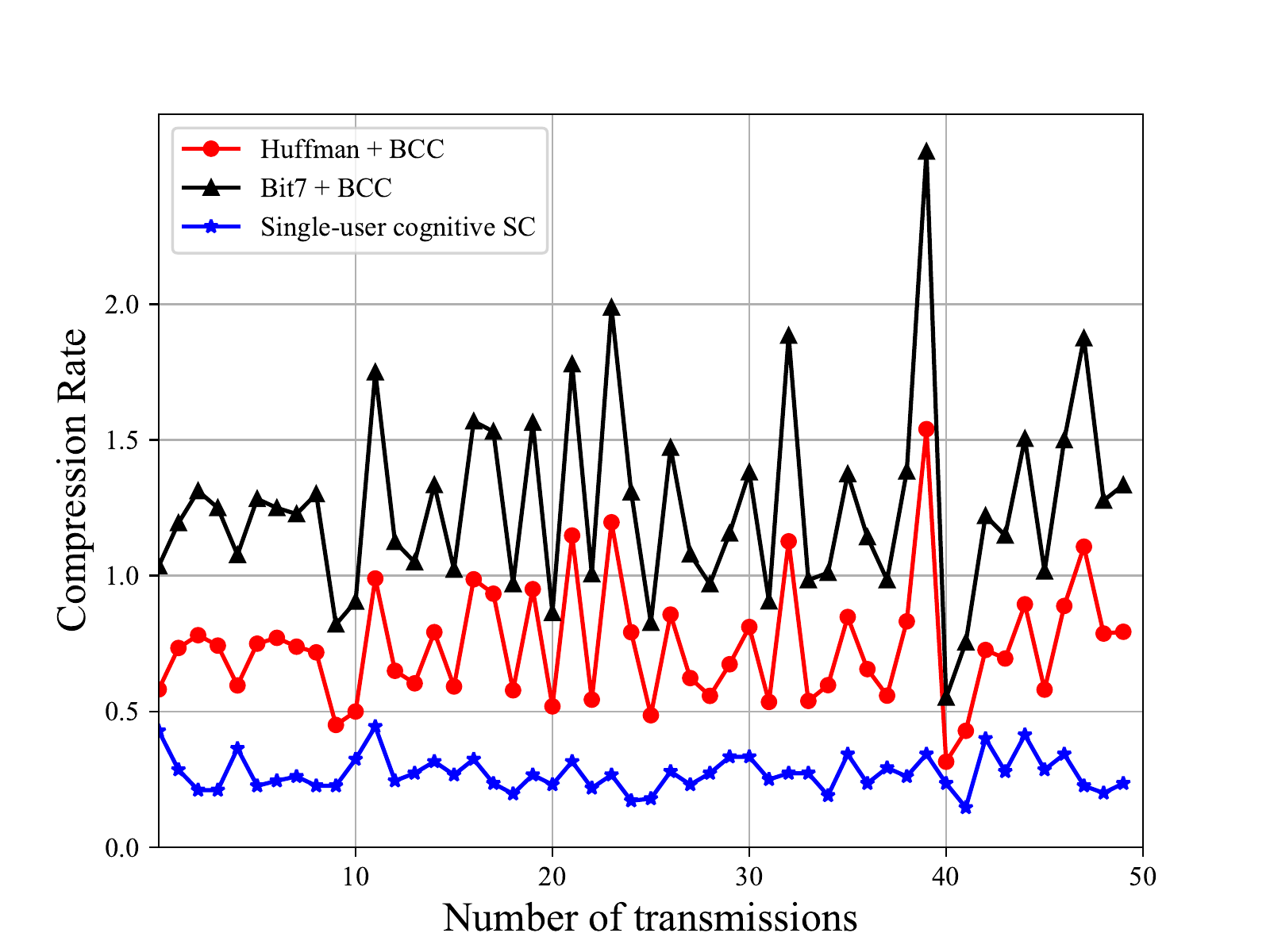}
    }
    \subfigure[] {
        \label{fig:b}
        \includegraphics[width=2.5 in]{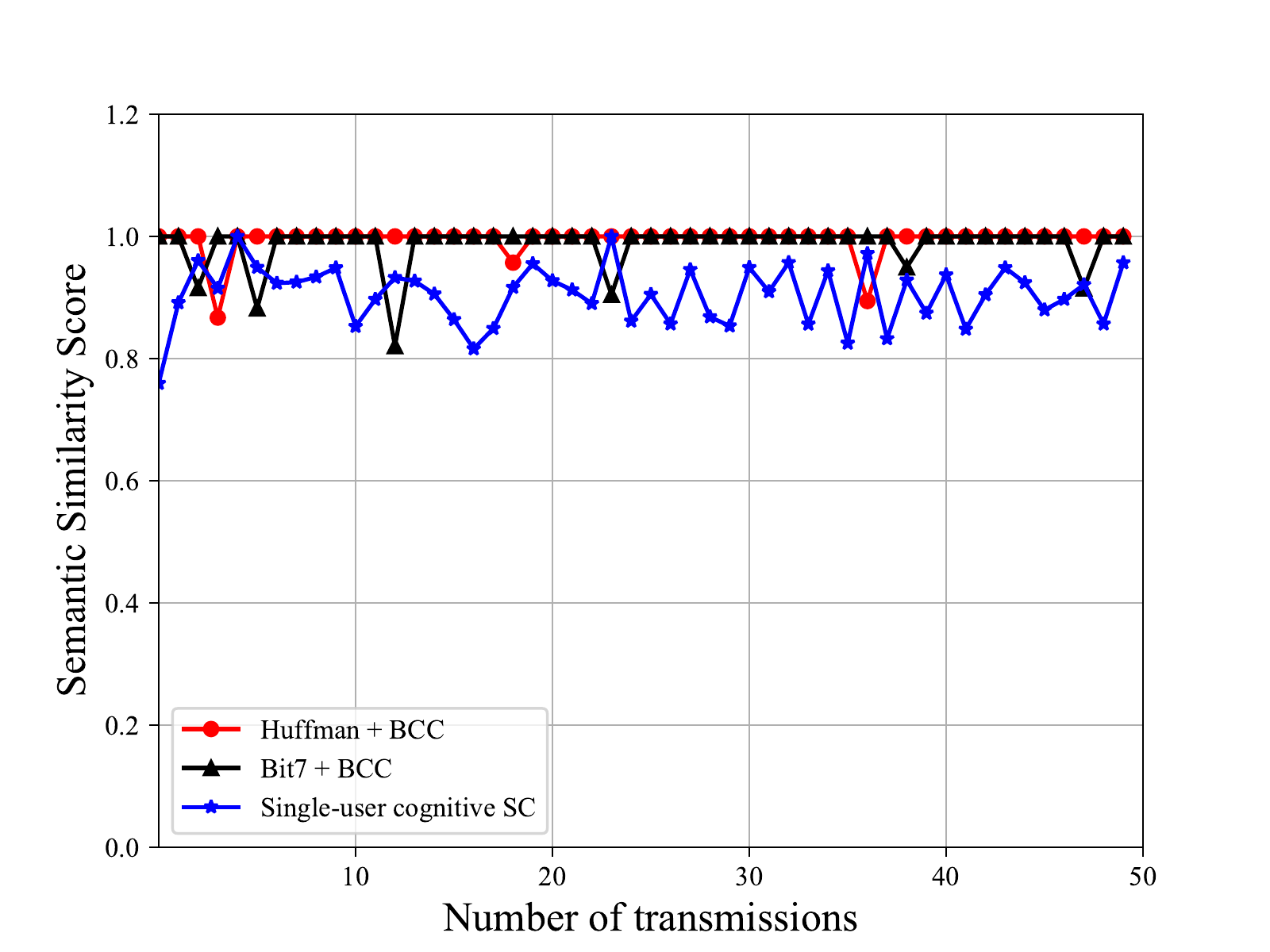}
    }
    \caption{Performance comparison by using different single-user schemes on our experimental SDR testbed.} 
    \label{fig:1}  
\end{figure}

\section{Conclusion}
In this paper, single-user and multi-user cognitive semantic communication systems were proposed by exploiting the knowledge graph. Moreover, a simple, general and interpretable semantic alignment algorithm for semantic information detection was proposed to improve communication effectiveness. In addition, a semantic correction algorithm was proposed by exploiting the inference rule of the knowledge graph to improve communication reliability. Furthermore, the T5 model was modified to recover semantic information in order to overcome the drawback that a fixed bit length coding is used to encode sentences with different lengths and results in low communication efficiency. For multi-user cognitive semantic communication system, a message recovery algorithm was proposed to distinguish messages of different users at the destination. The proposed single-user and multi-user cognitive semantic communication systems achieve promising performance. Simulation results demonstrated that the proposed systems have a superior data compression rate and communication reliability. The simulation results also demonstrated the effectiveness of our proposed semantic correction algorithm. Furthermore, the implementation on the USRP demonstrated the feasibility of our proposed cognitive semantic communication systems.

\bibliographystyle{IEEEtran}
\bibliography{ref}

\end{document}